\documentclass{article}
\usepackage{titling}
\usepackage{graphicx}
\usepackage{subcaption}
\usepackage{bm}
\usepackage{url}
\usepackage{fixltx2e}
\usepackage{multirow}
\usepackage{booktabs}
\usepackage{danudefs}
\usepackage[T1]{fontenc}
\usepackage[english]{babel}

\newtheorem{theorem}{Theorem}

\usepackage{tabularx,arydshln}
\usepackage{bm,amsfonts,amssymb,hhline}

\usepackage[sort]{natbib}

\usepackage{algorithmic}
\usepackage{algorithm}

\usepackage{graphicx}
\usepackage{epstopdf}

\renewcommand{\cite}{\citep}

\title{ClassiNet -- Predicting Missing Features for Short-Text Classification} 
 
\author{Danushka Bollegala \and
Vincent Atanaso \and Takanori Maehara \and Ken-ichi Kawarabayashi}

\date{}

\begin{document}
% Title portion

\maketitle

\begin{abstract}
Short and sparse texts such as tweets, search engine snippets, product reviews, chat messages are abundant on the Web.
Classifying such short-texts into a pre-defined set of categories is a common problem that arises in various contexts, such as
sentiment classification, spam detection, and information recommendation.
The fundamental problem in short-text classification is \emph{feature sparseness} -- the lack of feature overlap between
a trained model and a test instance to be classified. 
We propose \emph{ClassiNet} -- a network of classifiers trained for predicting missing features in a given instance, 
 to overcome the feature sparseness problem. 
Using a set of unlabeled training instances, we first learn binary classifiers as feature predictors
for predicting whether a particular feature occurs in a given instance.
Next, each feature predictor is represented as a vertex $v_i$ in the ClassiNet 
where a one-to-one correspondence exists between feature predictors and vertices.
The weight of the directed edge $e_{ij}$ connecting a vertex $v_i$ to a vertex $v_j$ represents 
the conditional probability that given $v_i$ exists in an instance, $v_j$ also exists in the same instance.

We show that ClassiNets generalize word co-occurrence graphs by considering implicit co-occurrences between features.
We extract numerous features from the trained ClassiNet to overcome feature sparseness.
In particular, for a given instance $\vec{x}$, we find similar features from ClassiNet that did not appear in $\vec{x}$, and append
those features in the representation of $\vec{x}$. Moreover, we propose a method based on graph propagation to find features that
are indirectly related to a given short-text.
We evaluate ClassiNets on several benchmark datasets for short-text classification.
Our experimental results show that by using ClassiNet, we can statistically significantly improve the accuracy in short-text classification tasks,
without having to use any external resources such as thesauri for finding related features.
\end{abstract}

\section{Introduction}
\label{sec:intro}

% growing amounts of short-texts on the web.
Short-texts are abundant on the Web and appear in various different formats. 
For example, in Twitter, users are constrained to
a $140$ character upper limit when posting their tweets~\cite{Kwak:WWW:2010}. 
Even when there are no strict upper limits, users tend to provide
brief answers in QA forums, review sites, SMS, email, and chat messages~\cite{Cong:SIGIR:2008,Thelwall:2010}.
Unlike lengthy responses that take time to both compose and to read, short responses have gained popularity particularly in 
social media contexts. Considering the steady growth of mobile devices that are physically restricted to compact keyboards,
which are suboptimal for entering lengthy text inputs, it is safe to predict that the amount of short-texts will continue to grow in the future.
Considering the importance and the quantity of the short-texts in various web-related tasks, such as
text classification~\cite{Wang:JZUS:2012,dossantos-gatti:2014:Coling},
 and event prediction~\cite{Sakaki:WWW:2010}, it is important to be able to accurately represent and classify short-texts.

% feature sparseness
Compared to performing text mining on longer texts~\cite{Yogatama:ICML:2014,Su:ICML:2011,Guan:WWW:2009}, for which dense and diverse
feature representations can be created relatively easily, handling of shorter texts poses several challenges.
First, the number of features that are actually present in a short-text will be a small fraction of the set of all features that exist
in all of the train instances. Although this \emph{feature sparseness} is problematic even for longer texts, it is critical for shorter texts.
In particular, when the diversity of the feature space increases as with longer $n$-gram lexical features, 
(a) the number of occurrences of a feature in a given instance (i.e., term frequency), as well as 
(b) the number of instances in which a particular feature occurs (i.e., document frequency),
will be small. Therefore, it is difficult to reliably estimate the salience of a feature in a particular class in
supervised learning tasks.

% challenges
Second, the shorter length means that there is \emph{less redundancy} in terms of the features that exist in a short-text. Consequently,
most of the related words of a particular word might be missing in a short-text. 
For example, consider a review on \emph{iPhone 6} that says ``\emph{I liked the larger screen size of iPhone 6 compared to
that of its predecessor}''. Although \emph{iPhone 6 plus}, a product similar to \emph{iPhone 6}, has also a larger screen compared to
its predecessors, this information is not included in this short review. On the other hand, we might observe such positive sentiments
associated with \emph{iPhone 6 plus} but not with \emph{iPhone 6} in other train instances,
 which will result in a high positive score for \emph{iPhone 6 plus} in
a classifier trained from those train reviews. Unfortunately, we will not be able to infer that this particular user would also likely
be satisfied with \emph{iPhone 6 plus}, thereby not recommending \emph{iPhone 6 plus} for this user.

% solution
To overcome the above-mentioned challenges encountered when handling short-texts, we propose a \emph{feature expansion} method
analogous to the query expansion methods used in information retrieval (IR)~\cite{IR_book} to improve 
the agreement between search queries input by the users and documents indexed by the search engine~\cite{Carpineto:2012}.
We assume short-texts are already represented using some feature vectors, which we refer to as \emph{instances}
in this paper. Lexical features such as unigrams or bigrams of words, part-of-speech (POS) tag sequences, and dependency relations
have been frequently used in prior work on text classification.
Our proposed method does not assume any particular type of features, and can be used with any discrete feature set.
First, we train binary classifiers which we call \emph{feature predictors} for predicting whether a particular feature $v_i$ occurs
in a given instance $\vec{x}$. 
For example, given the previously discussed short review, 
we would like to predict whether iPhone 6 plus is likely to occur in this review. 

The training instances required to learn feature predictors are automatically selected from unlabeled texts.
Specifically, given a feature $v_i$, we select texts in which $v_i$ occurs as the positive training instances for
 learning a feature predictor for $v_i$.
On the other hand, negative training instances for learning the feature predictor for $v_i$ are randomly sampled from the
unlabeled texts, where $v_i$ does not occur. 
Using those positive and negative training instances we learn a binary classifier to predict whether $v_i$ occurs in a given instance.
Any binary classification algorithm, such as support vector machines, logistic regression, naive Bayes classifier etc. 
can be used for this purpose, and it is not limited to linear classifiers.
We define \emph{ClassiNet} as a directed weighted graph $\cG(\cV, \cE, \mat{W})$ 
of feature predictors, where each vertex $v_i \in \cV$ corresponds to a feature predictor. 
The directed edge $e_{ij} \in \cE$ from $v_i$ to $v_j$ is assigned the weight $1 \geq w_{ij} \geq 0$, which is the
conditional probability that given $v_i$ is predicted for a particular instance, $v_j$ is also predicted for the same instance.

It is noteworthy that we obtain both positive and negative instances for learning feature predictors from
unlabeled data, and do not require any labeled data for the target task.
For example, consider the case that we are creating a ClassiNet to find missing features in sentiment classification.
In this case, the target task is sentiment classification. However, we do not require any labeled data for the target task
such as sentiment annotated reviews when creating the ClassiNet that we are subsequently going to use for finding missing features.
Therefore, the training of ClassiNets can be conducted in a purely unsupervised manner,
without requiring any manually labeled data for the target task. 
Moreover, the decoupling of ClassiNet training from the target task enables us to
use the same ClassiNet to expand feature vectors for different target tasks.
As we discuss later in Section~\ref{sec:classi-cooc}, ClassiNets can be seen as a generalized version
 of the word co-occurrence graphs that have been
well-studied in the NLP community~\cite{Rada:2011}. 
However, ClassiNets consider both explicit as well as implicit co-occurrences of words in some context,
whereas word co-occurrence graphs are limited to explicit co-occurrences.

% contribution
Given a ClassiNet created from unlabeled data as described above, we propose several strategies for finding 
related features for a given instance that do not occur in the original instance. 
Specifically, we compare both \emph{local} feature expansion methods that consider the nearest neighbours of a particular
feature in an instance (Section~\ref{sec:local}), 
as well as \emph{global} feature expansion methods that propagate the features that exist in an instance
over the entire set of vertices in ClassiNet (Section~\ref{sec:global}). 
We evaluate the performance of the proposed feature expansion methods
on short-text classification benchmark datasets.
Our experimental results show that the proposed global feature expansion method 
significantly outperforms several local feature expansion methods,, and several sentence-level
 embedding methods on multiple benchmark datasets proposed for evaluating short-text classification methods. 
%Interestingly, the optimal damping factor for the global feature expansion turns out to be $0.85$, which is known to work well
%for the PageRank algorithm on Web link graphs~\cite{PageRank}.
Considering that (a) ClassiNets can be created using unlabeled data, 
(b) the same ClassiNet can be used in principle for predicting features for different target tasks, 
(c) arbitrary features could be used in the feature predictors, not limited to lexical features,
we believe that ClassiNets can be applied to a broad-range of machine learning tasks, not limited to short-text classification.

Our contributions in this paper can be summarised as follows:
\begin{itemize}
\item We propose a method for learning a network of feature predictors that can predict missing features in feature vectors.
The proposed network, which we refer to as the ClassiNet, can be learnt in an unsupervised manner, without requiring any labeled
data for the target task in which we are going to apply the ClassiNet to expand features (Section~\ref{sec:classinet:learn}).

\item We propose an efficient method to learn ClassiNets from large datasets. Specifically, we show that the edge-weights
of ClassiNets can be computed efficiently using locality sensitive hashing (Section~\ref{sec:project}).

\item Having proposed ClassiNets, we describe its relationship to word co-occurrence graphs that have a long history in
the NLP community. We show that ClassiNets can be considered as a generalised version of
word co-occurrence graphs (Section~\ref{sec:classi-cooc}).

\item We propose several methods for finding related features for a given instance using the created ClassiNet. In particular,
we consider both \emph{local methods} (Section~\ref{sec:local}) that consider the nearest neighbours
 in ClassiNet of the features that exist in an instance,
as well as \emph{global methods} (Section~\ref{sec:global}) that consider all vertices in the ClassiNet.
\end{itemize}

\section{Related Work}
\label{sec:related}

% Two approaches: extend or dimensionality reduction
% expansion using WordNet (not good as we show later)

% A word on co-occurrence graphs and the classinet
% SCL, SFA domain adaptation

% text similarity
% \cite{Metzler:2007}
% Sahami \cite{Sahami:WWW:2006}

% Short text classification

% tweet normalization \cite{BHan:ACL:2011}

% Sentiment classification of short texts such as sentences \cite{Thelwall:2010} or tweets.
% Dimensionality reduction by LDA \cite{Wang:JZUS:2012}
% Phrase topic model \cite{yang-EtAl:2015:NAACL-HLT2}
% Bi-term topic model \cite{Yan:WWW:2013}
% Cluster-based representation \cite{Dai:2013}
% Clustering of short texts \cite{Rangrej:WWW:2011}

% CNN \cite{dossantos-gatti:2014:Coling}
% Deep architecture for matching short texts \cite{Lu:NIPS:2013}

% Frequent itemsets \cite{Man:2014}
% Hungarian \cite{song-roth:2015:NAACL-HLT}

Feature sparseness is a common problem that is encountered in various text mining tasks.
Two main approaches for overcoming the feature sparseness problem in short-texts can be identified in the literature:
(a) embedding the train/test instances in a dense, lower-dimensional feature space 
thereby reducing the number of zero-valued features in the instances, and
(b) predicting the values of the missing features. 
Next, we discuss prior work that belong to each of those two approaches.

An effective technique frequently used in prior work on short-texts to overcome the feature sparseness problem
 is to represent the texts in some lower-dimensional dense space,
thereby reducing the feature sparseness. Several methods have been used to obtain such lower-dimensional representations such
as topic-models~\cite{Yan:WWW:2013,yang-EtAl:2015:NAACL-HLT2,Wang:JZUS:2012},
clustering~\cite{Dai:2013,Rangrej:WWW:2011}, and dimensionality reduction~\cite{Blitzer:EMNLP:2006,Pan:WWW:2010}.
Wang et al.~\cite{Wang:JZUS:2012} used latent dirichlet allocation (LDA) to identify features that are useful for identifying a particular
class. Higher weights are assigned to the identified features, thereby increasing their contribution towards the classification decision.
However, applying LDA at sentence-level is problematic because the number of words in a sentence is much smaller than that in a 
document. Consequently, Yan et al.~\cite{Yan:WWW:2013} proposed the bi-term topic model that models the co-occurrence patterns
between words accumulated over the entire corpus. 
An alternative solution that uses an external knowledge-base in the form of a phrase list is propsed by 
Yang et al.~\cite{yang-EtAl:2015:NAACL-HLT2} to overcome the feature sparseness problem when learning topics from short-texts.
The phrase list is automatically extracted from the entire collection of short-texts in a pre-processing step.

Cluster-based methods have been proposed for representing documents to overcome the feature sparseness problem.
First, some clustering algorithm is used to cluster the documents into a group of clusters.
Next, each document is represented by the clusters to which it belongs.
Dai et al.~\cite{Dai:2013} used a hierarchical clustering algorithm with purity control to generate a set of clusters,
and use the similarity between a document and each of the clusters as augmented features to enrich the document representation.
Their method significantly improves the classification accuracy for short web snippets in a support vector machine classifier.
Feature mismatch is a fundamental problem in domain adaptation, where we must learn a classifier using labeled data from
a source domain and apply it to predict labels for the test instances in a different target domain.
Pan et al.~\cite{Pan:WWW:2010} proposed Spectral Feature Alignment (SFA),
 a method to overcome the feature mismatch problem in cross-domain sentiment
classification. They created a bi-partite graph between domain-specific and domain-independent features,
and then used a spectral clustering method to obtain a domain-independent lower-dimensional embedding.

In structural correspondence learning (SCL)~\cite{Blitzer:ACL:2007,Blitzer:EMNLP:2006},
a set of features that are common to both source and the target domains, referred to as \emph{pivots},
is identified using mutual information with the sentiment label. Next, linear classifiers that can predict those pivots are learnt
from unlabeled reviews. The weight vectors corresponding to the learnt linear classifiers are arranged as rows in a matrix,
on which subsequently singular value decomposition is applied to compute a lower-dimensional projection.
Feature vectors representing train source reviews are projected into this lower-dimensional space, 
in which a binary sentiment classifier is trained.
During test time, feature vectors representing test target reviews are also projected to the same lower-dimensional space
and the trained binary classifier is used to predict the sentiment labels.
However, domain adaptation methods such as SCL and SFA require data from at least two (source vs. target) different domains
(e.g. reviews on products in different categories) to overcome the missing feature problem, whereas in this work we
assume the availability of data from one domain only.

Instead of representing documents using lexical features, which often results in high-dimensional and sparse feature vectors,
by embedding documents in low-dimensional dense spaces 
we can effectively overcome the feature sparseness problem~\cite{Lu:NIPS:2013,dossantos-gatti:2014:Coling,Le:ICML:2014}. 
These methods jointly learn character-level or word-level embeddings as well as 
document-level embeddings~\cite{Kiros:2015,Hill:NAACL:2016} such that
the learnt embeddings capture the similarity constraints satisfied by a collection of short-texts.
First, each word in the vocabulary is assigned a fixed dimensional word vector. We can initialize the word vectors randomly or
using pre-trained word representations. Next, the word vectors are updated such that we can accurately
predict the co-occurrences of words in some context, such as a window of tokens, a sentence, a paragraph, or a document.
Different loss functions encoding different co-occurrence measures
 have been proposed for this purpose~\cite{Pennington:EMNLP:2014,Milkov:2013}.
 As shown later in Section~\ref{sec:sentemb}, ClassiNets perform competitively against sentence-level embedding methods
 on several short-text classification tasks.
 
A single word can have multiple senses. For example, the word \emph{bank} could mean a \emph{financial institution} or a \emph{river bank}.
Therefore, it is inadequate to represent different senses of a word using a single embedding~\cite{Reisinger:NAACL:2010,Iacobacci:ACL,Song:2016,camachocollados-pilehvar-navigli:2015:NAACL-HLT,johansson-nietopina:2015:NAACL-HLT,li-jurafsky:2015:EMNLP,hu-zhang-zheng:2016:COLING}.
Several solutions have been proposed in the literature to overcome this limitation and learn \emph{sense embeddings}, which capture the sense related information of words. For example, \citet{Reisinger:NAACL:2010} proposed a method for learning sense-specific high dimensional distributional vector representations of words, which was later extended by \citet{Huang:ACL:2012} using global and local context to learn multiple sense embeddings for an ambiguous word.
\citet{neelakantan-EtAl:2014:EMNLP2014} proposed a multi sense skip-gram (MSSG), an online cluster-based sense-specific word representations learning method, by extending Skip-Gram with Negative Sampling (SGNG)~\cite{Milkov:2013}.
Unlike SGNG, which updates the gradient of the word vector according to the context, MSSG predicts the nearest sense first, and then updates the gradient of the sense vector.

Aforementioned methods apply a form of word sense discrimination by clustering a word contexts, before learning sense-specific word embeddings based on the induced clusters to learn a fixed number of sense embeddings for each word.
In contrast, a nonparametric version of MSSG (NP-MSSG)~\cite{neelakantan-EtAl:2014:EMNLP2014} estimates the number of senses per word
and learn the corresponding sense embeddings.
On the other hand, \citet{iacobacci-pilehvar-navigli:2015:ACL-IJCNLP} used a Word Sense Disambiguation (WSD) tool to sense annotate a large text corpus and then used an existing prediction-based word embeddings learning method to learn sense and word embeddings with the help of sense information obtained from the BabelNet~\cite{iacobacci-pilehvar-navigli:2015:ACL-IJCNLP} sense inventory.
Similarly, \citet{camachocollados-pilehvar-navigli:2015:NAACL-HLT} used the knowledge in two different lexical resources: WordNet~\cite{WordNet} and Wikipedia. 
They use the contextual information of a particular concept from Wikipedia and WordNet synsets prior to learning two separate vector representations for each concept.

A single word can be related to multiple different topics, without necessarily corresponding to different senses of the word. 
Revisiting our previous example, we might have a collection of documents about \emph{retail banks}, \emph{commercial banks}, \emph{investment banks} and \emph{central banks}. All these different banks are related to the financial sense of the word bank.
However, in a particular task (eg. classifying documents related to the different types of financial banks), we might require different embeddings for the different topics in which the word bank appears. 
\citet{Liu:AAAI:2015} proposed three methods for learning \emph{topical word embeddings}, where they first cluster words into different topics using LDA~\cite{Blei:JMLR:2003} and then learn word embeddings using SGNS.
\citet{Liu:IJCAI:2015} modelled the interactions among topics, contexts and words using a tensor and obtained topical word embeddings via tensor factorisation.
Instead of clustering words prior to embedding learning, \citet{Shi:2017} proposed a method to jointly learn both words and topics, thereby considering the correlations between multiple senses of different words that occur in different topics.
TopicVec~\cite{TopicVec} learns vector representations for topics in a document by modelling the co-occurrence between a target word and a context word considering both words' word embeddings as well as the topic embedding of the context word.

Our proposed methods for feature expansion using ClassiNet can be seen as an \emph{explicit} feature prediction method,
whereas methods that learn lower-dimensional dense embeddings of texts can be seen as \emph{implicit} feature prediction methods.
For example, if we use lexical features such as unigrams or bigrams to create a ClassiNet, then the features predicted by that ClassiNet
will also be lexicalised features, which are easier to interpret than dimensions in a latent embedded space.
Although for text classification purposes it is sufficient to represent short-texts in implicit feature spaces, there are numerous
tasks that require explicit interpretable predictions such as
query suggestion in information retrieval~\cite{Carpineto:2012}, 
reverse dictionary mapping~\cite{Hill:TACL:2016}, and hashtag suggestion in social media~\cite{weston-chopra-adams:2014:EMNLP2014}.
Therefore, the potential applications of ClassiNets as an explicit feature expansion method goes beyond short-text classificaion.
It would be an interesting future research direction to combine implicit and explicit feature expansion methods to
construct better representations for texts.

Recently there has been several methods proposed for learning embeddings (lower-dimensional implicit feature representations) for the vertices
of undirected or directed (and weighted) graphs~\cite{DeepWalk,li-zhu-zhang:2016:P16-1,LINE}. 
For example, in \emph{language graphs}~\cite{LINE}, the vertices can correspond to words and the
weight of the edge between two vertices represent the strength of the co-occurrences between two words in a corpus.
Alternatively, in a \emph{co-author network}, the vertices correspond to authors and the edges represent the number of papers two people have
co-authored.  DeepWalk~\cite{DeepWalk} performs a random walk over an undirected graph to generate a pseudo-corpus, which is then used to learn word (vertex) embeddings using skip-gram with negative sampling (SGNS)~\cite{Milkov:2013}.
Li et al.~\cite{li-zhu-zhang:2016:P16-1} proposed a discriminative version of DeepWalk by including a discriminative supervised loss that evaluates how well the learnt vertex embeddings perform on some supervised tasks.
Tang et al.~\cite{LINE} used both first-order and second-order co-occurrences in a graph to learn separate vertex embeddings, which were subsequently concatenated to create a single vertex embedding.
Although in this paper we consider graphs where vertices correspond to words, the objective of creating ClassiNets is fundamentally different
from the above-mentioned vertex embedding methods.
In graph (vertex) embedding, we are given a graph and a goal is to learn embeddings for the vertices such that structural information of the graph
is preserved in the learnt embeddings. On the other hand, in ClassiNets, we learn feature predictors which can be used to predict whether a particular
feature is missing in a given context. The connection between co-occurrence graphs and ClassiNets is further discussed in Section~\ref{sec:classi-cooc}. Moreover, in Section~\ref{sec:classinet:expand}, we propose and evaluate several methods for expanding feature vectors using the ClassiNets we create, which is not relevant 
for vertex embedding methods.

\section{ClassiNets}
\label{sec:classinets}

\subsection{Overview}
\label{sec:overview}

Our proposed method for classifying short-texts consists of two steps. First, we create a network of classifiers which we refer to
as the \emph{ClassiNet} in this paper. In Section~\ref{sec:classinet:learn}, we describe the details of the method we propose to create
ClassiNets. In Section~\ref{sec:classinet:expand}, we describe several methods for using the learnt ClassiNet to expand feature vectors
to overcome the feature sparseness problem.

We define a ClassiNet as a directed weighted graph $\cG(\cV, \cE, \mat{W})$, in which a vertex $v_i \in \cV = \{v_1, \ldots, v_n \}$ 
corresponds to a binary classifier (feature predictor) $h_i$ that predicts the occurrence of a feature $v_i$ in an instance.
We assume that each train/test instance $x$ is already represented by a $d$-dimensional
vector $\vec{x} = (x_1, x_2, \ldots, x_d)\T$, in which the $i$-th dimension corresponds to the value $x_i$ of the $i$-th feature
representing the instance $x$.
The label predicted by $h_i$ for an instance $\vec{x}$ is denoted by $h_i(\vec{x}) \in \{0,1\}$.
The weight $w_{ij}$ associated with the edge $e_{ij}$ connecting the vertex $v_i$ to $v_j$ 
represents the conditional probability, $p(h_j(\vec{x}) = 1| h_i(\vec{x}) = 1)$,  
that $v_j$ is predicted to occur in $\vec{x}$, given that $v_i$ is also predicted to occur in $x$.

Several remarks can be made about the ClassiNets.
First, there is a one-to-one correspondence between the vertices $v_i$ in the ClassiNet and the feature predictors $h_i$.
Therefore, a ClassiNet can be seen as a network of binary classifiers, as is implied by its name.
In general, the set of features $\cS$ that we use for representing instances $x$ (hence for learning feature predictors), 
and the set of vertices $\cV$ in ClassiNet need not be the same. As we discuss later, vertices in the ClassiNet are used
as expansion features to augment instances $x$, thereby overcoming the feature sparseness problem in short-text classification.
Therefore, we are free to select a subset of features from all the features used for representing instances as the vertices in ClassiNet.
For example, we might use the most frequent features in the train data
as vertices in ClassiNet thereby setting $\cV \subset \cS$ ($n < d$).
Alternatively, we could use all the features in the feature space of the instances as vertices in the ClassiNet, where we have
$\cV = \cS$ (and $n = d$). In the remainder of the paper, we consider the general 
case where we have $\cV \subseteq \cS$ ($n \leq d$).

Second, as we discuss later in Section~\ref{sec:classinet:learn}, 
we \emph{do not} require labeled data for the target task when creating ClassiNets.
For example, let us consider binary sentiment classification of product reviews as the target task.
We might have both sentiment rated reviews (labeled instances), and reviews without sentiment ratings
(unlabeled instances) at our disposal.
We can use both those types of reviews, and ignore the label information when computing the ClassiNet.
This is particularly attractive for two reasons: 
(a) obtaining unlabeled instances is often easier for most tasks compared to obtaining labeled instances,
(b) because a ClassiNet created from a particular corpus is independent of the label information unique to a target task, 
in principle, the same ClassiNet can be used to expand features for different target tasks.
The second property is attractive in multi-task learning settings, where we must perform different tasks on the same data. 
For example, consider the two tasks:
(a) predicting whether a given tweet is positive or negative in sentiment, and 
(b) predicting whether a given tweet would get favorited or not.
Both those tasks can be seen as binary classification tasks. We could learn two binary classifiers -- one for predicting the sentiment
and the other for predicting whether a tweet would get favorited. 
However, to overcome the feature sparseness problem in both those tasks, we can use the same ClassiNet.

As long as an instance (for example a sentence or a document) is represented using any bag-of-features (unigrams, bigrams, trigrams, dependency paths, syntactic paths, POS sequences, semantic roles, frames etc.) we can use the proposed method to create a ClassiNet. The first step in creating a ClassiNet is to learn feature predictors (Section~\ref{sec:classinet:learn}). The feature predictors use the features available in an instance to as features to train a binary classifier. Therefore, it does not matter whether these features are $n$-grams or more complex types of features as listed above. The remainder of the steps in the proposed method (measuring the correlations between feature predictors to build the ClassiNet, applying feature expansion) use only the learnt feature predictors. Therefore, our proposed method can be used with \emph{any} feature representation of instances, not limiting to lexical n-gram features.

%By definition the edge-weights $w_{ij}$ ($0 \leq w_{ij} \leq 1$) in ClassiNets are asymmetric (i.e., $w_{ij} \neq w_{ji}$). 
%Moreover, from the properties of conditional probabilities we have, 
%\[
%\sum_{j=1}^{n} w_{ij} = \sum_{j=1}^{n} p(h_j(\vec{x}) = 1|h_i(\vec{x}) = 1) = p(h_i(\vec{x}) = 1)
%\]
% satisfied at each vertex of the ClassiNet.
%Moreover, the sum of all the edge-weights in ClassiNet equals to one ($\sum_{i=1}^{n} \sum_{j=1}^{n} w_{ij} = 1$).

\subsection{Learning ClassiNets}
\label{sec:classinet:learn}

Let us assume that we are given a set $\cD_{u} = \{\vec{x}^{(k)}\}_{k=1}^{N}$ of unlabeled feature vectors $\vec{x}^{(k)} \in \R^d$
 representing $N$ short-texts.
Given $\cD_{u}$ we construct a ClassiNet in two steps: (a) learn feature predictors $h_i$ for each vertex $v_i \in \cV$,
and (b) compute the conditional probabilities $p(h_j(\vec{x}) = 1| h_i(\vec{x}) = 1)$ 
using the labels predicted by the feature predictors $h_i$ and $h_j$ for an instance $\vec{x}$.
As positive training instances for learning a binary feature predictor for a feature $v_i$, 
we randomly select a set $\cD_i^{(+)} \subset \cD_{u}$ of $N^{(+)}_i$ instances where $v_i$ occurs,
and remove $v_i$ from those selected instances.
% (i.e. $\cD_i^{(+)} = \{\vec{x}| \vec{x} \in \cD_{u}, x_i \neq 0\}$).
Likewise, we randomly select a set $\cD_i^{(-)} \subset \cD_{u}$ of $N^{(-)}_i$ instances where $v_i$ does not occur.
Instances that have few features are not informative for learning accurate feature predictors.
Therefore, we select instances that have more non-zero features than the average number of non-zero 
features in an instance in $\cD_{u}$. We found that, on average, there are ca. $15$ features in an instance.

Compared to the number of instances containing a particular feature $v_i$ in the dataset, 
the number of instances that do not contain $v_i$ is significantly larger.
Considering that we are randomly sampling negative instances from a larger set of instances,
it is likely that those selected negative instances are not very informative about why $v_i$ is missing in a given instance.
In other words, the randomly sampled negative instances might already be further from the decision hyperplane, therefore
do not provide sufficient specialization in the hypothesis space. 
Consequently, it has shown in prior work that use pseudo-negative instances for training classifiers~\cite{Bollegala_WWW_2007} that it is effective to
select a larger number of pseudo-negative instances than that of positive instances (i.e., $N^{(+)}_i < N^{(-)}_i$).
We note that it is possible to set the number of positive and negative train instances
dynamically for each feature $v_i$. For example, some features might be popular in the dataset resulting in a larger positive sample
than the others. For simplicity, in this paper, we select all instances in which a particular feature occurs as 
the positive training instances for that feature, and select twice that number of negative instances from the remainder of the instances
(i.e., $N^{(-)}_i = 2N^{(+)}$). 
An extensive study of different sampling methods and  $N^{(-)}_i / N^{(+)}_i$ ratios is beyond the scope of the current paper.

Once we have selected $\cD_i^{(+)}$, and $\cD_i^{(-)}$ as described above, we train a binary classifier to predict whether
$v_i$ occurs in a given instance. We note that any binary classification algorithm, not limited to linear classifiers,
can be used for this purpose. In our experiments, we use $\ell_2$ regularized logistic regression for its simplicity. 
We tune the regularization coefficient in each feature predictor using $5$-fold cross-validation.
Being a probabilistic discriminative classifier, it is possible to obtain not only the predicted labels
but also the class conditional probabilities from the trained logistic regression classifier. 
However, we only require the predicted labels for constructing the edge weights in ClassiNets
as we describe next. Therefore, in theory, we can use even binary classifiers that do not produce confidence scores
for creating ClassiNets, which extends the applicability of ClassiNets to wider contexts.

Let us denote the label predicted by the feature predictor $h_i$ for an instance $\vec{x}$ by $h_i(\vec{x}) \in \{0,1\}$.
For two features $v_i$ and $v_j$, we compute the confusion matrix $\mat{M}$ shown in Table~\ref{tbl:conf}.
Here, $M_{ab}$ denotes the number of instances $\vec{x}$ for which
$h_i(\vec{x}) = a$ and $h_j(\vec{x}) = b$. 
In particular, $M_{11}$ is the number of instances where both $v_i$ and $v_j$ are predicted 
to be co-occurring by the learnt feature predictors.

\begin{table}[t]
\centering
\caption{Confusion matrix for the labels predicted by the feature predictors learnt for two features $v_i$ and $v_j$.}
\label{tbl:conf}
\begin{tabular}{|c|c|c|}\hline
				& $h_j(\vec{x}) = 1$		&	$h_j(\vec{x}) = 0$ \\ \hline
$h_i(\vec{x}) = 1$	&	$M_{11}$			&	$M_{10}$ \\ \hline
$h_i(\vec{x}) = 0$	&	$M_{01}$			& 	$M_{00}$ \\ \hline
\end{tabular}
\end{table}

Given the counts in Table~\ref{tbl:conf}, $w_{ij}$ is computed as follows:
\begin{equation}
\label{eq:weight} 
w_{ij} = \frac{M_{11}}{M_{11} + M_{10}}
\end{equation}
Several practical issues must be considered when estimating the edge-weights using \eqref{eq:weight}.
First, the set of instances we use for predicting labels when computing the confusion matrix in Table~\ref{tbl:conf} must contain
at least some instances in which $v_i$ or $v_j$ occur (i.e., $M_{11} + M_{10} > 0$, and $M_{11} + M_{01} > 0$). 
Otherwise, even if the feature predictors $h_i$, $h_j$ are accurately learnt, 
we will still get unreliable sparse counts for $M_{11}$ and $M_{10}$.
Therefore, we randomly sample a set of instances $\cD_{(i,j)} \subseteq \cD_{u}$ such that 
there exist equal numbers of instances containing $v_i$, and $v_j$.

Let the total number of elements in $\cD_{(i,j)}$ be $d'$. 
We use those $d'$ instances when computing the values in the confusion matrix shown in Table~\ref{tbl:conf}. 
We ensure that there is no overlap between the test instances $\cD_{(i,j)}$ and the train instances we use to learn
feature predictors. This is important because if the feature predictors are overfitting we will not get accurate predictions using
the ClassiNet during test time. Using non-overlapping train and test instance sets, we can check whether the learnt
feature predictors are overfitting. Although we use a ratio of one-third when sampling $\cD_{(i,j)}$ above,
we can use different ratios for sampling as long as both $v_i$ and $v_j$ are sufficiently represented in $\cD_{(i,j)}$.

\subsection{Efficient Computation of ClassiNets}
\label{sec:project}

ClassiNets can be learnt offline during the training stage, prior to expanding test instances. 
Therefore, we are allowed to perform more computationally intensive
processing steps compared to what we are allowed at test time, which is required to be real-time for most tasks that involve short-texts
such as tweet classification. 
Nevertheless, we propose several methods to
speed-up the the construction process when the number of vertices $n$ in the ClassiNet grows.

Compared to learning feature predictors for the vertices we use in the ClassiNet, which is linear in the number of
vertices $n$ in the ClassiNet, to compute weights $w_{ij}$ we must consider all pairwise combinations  between the vertices in the ClassiNet.
If we assume that the cost of learning a binary classifier for a vertex to be a constant $c$ and is independent of the feature, then
the overall computational complexity of creating a ClassiNet can be estimated as $\cO(cn + N n^2 d )$.
The first term is simply the complexity of computing $n$ feature predictors at the constant cost of $c$.
This operation can be easily parallelised because each feature predictor can be learnt independently
of the others.  Moreover, it is linear in the number of vertices in ClassiNet.
Therefore, the first term can be ignored in most practical scenarios.

In cases where computational cost of the linear predictors is non-negligible, we can use several techniques to speed up this computation.
First, we could resort to more computationally efficient liner classifiers such as the perceptron. Perceptrons can be trained in an online manner,
without having to load the entire training dataset to the memory.
Second, note that only the features $v_{j}$ that co-occur with a particular vertex $v_{i}$ in any train instance will be useful for predicting the occurrence of $v_{i}$. Therefore, we can limit the features that we use in the predictor for $v_{i}$ to be the set of features $v_{j}$ that occur at least once in the training data. We can efficiently compute such feature co-occurrences by building an inverted search index.
We can further speed up this computation by resorting to approximate methods where we require a context feature $v_{j}$ to co-occur a predefined minimum number of times with the target feature $v_{i}$ for which we must compute a predictor. 
Setting this cut-off threshold to higher values will result in smaller, sparser and less noisier feature spaces and speed up the predictor computation.
However, larger cut-off thresholds are likely to remove important contextual features, thereby decreasing the accuracy of the feature predictors.
The optimal cut-off threshold could be determined using cross-validation or held-out data.

On the other hand, the second term corresponds to learning edge-weights, and involves three factors: 
(a) $n^2$, the number of pairwise comparisons we must perform between the $n$ vertices in the ClassiNet, 
(b) $N$,  the maximum number of instances for which we must predict labels for each pair of feature predictors
when we compute the confusion matrices as shown in Table~\ref{tbl:conf}, and
(c) $d$, the number of features we must consider when computing the label of a predictor. For example, if we use
linear classifiers as feature predictors, during test time we must compute the inner-product between the 
weight vector of the classifier and the feature vector of the instance to be classified, both of which are $d$-dimensional.
The dimensionality $d$ of the vectors that represent instances will depend on the type of features we use.
For example, if we limit to lexical features from the short-text, then the number of non-zero
features in any given instance will be small. However, if we use dense features such as word embeddings,
then the number of non-zero features in an instance might be large.

However, the factors (a) and (b) require careful consideration.
First, we must compare all pairs of predictors, which is quadratic in the number of vertices in the ClassiNet.
Second, to obtain the label for an instance we must classify that instance using the learnt prediction model.
For example, in the case of linear classifiers we must compute the inner-product between two $d$-dimensional vectors:
feature vector representing the instance to be classified, and the weight vector corresponding to the feature predictor.
For nonliner classifiers such as the ones that use polynomial kernels, the number of feature combinations can grow
exponentially resulting in slower prediction times for large batches of test instances.

As a solution to this problem, we first represent each feature predictor $h_i$ by a $d' (< d)$ dimensional vector $\vec{h}_i(\cD_{(i,j)})$,
where each element corresponds to the label predicted for a particular instance $\vec{x} \in \cD_{(i,j)}$.
We randomly sample $\cD_{(i,j)} \subseteq \cD_{u}$ following the procedure detailed in Section~\ref{sec:classinet:learn},
where we include equal numbers of instances that contain $v_i$, $v_j$, and neither of those two.
Therefore,  $\vec{h}_i(\cD_{(i,j)}) \in \mat{I}_{d'}$
% where $d' = |\cD_{(i,j)}|$ is the cardinality of the subset we select from $\cD_{u}$,
and $\mat{I}_{d'}$ is the $d'$-dimensional simplex.
We name $\vec{h}_i(\cD_{(i,j)})$ as the \emph{label vector} because it is a vector of predicted labels 
for all the instances in $\cD_{(i,j)}$ by $h_i$, the feature predictor learnt for the feature $v_i$.
We can explicitly compute the label vector for the $i$-th feature predictor as follows:
\begin{equation}
\label{eq:label-vector}
\vec{h}_i(\cD_{(i,j)}) = \left( \vec{h}_i(\vec{x}_1), \ldots, \vec{h}_i(\vec{x}_{d'}) \right)\T
\end{equation}

In practice, $d' \ll N$ because only a small number of instances in $\cD_{u}$ will 
contain $v_i$, or $v_j$, and we select equal proportions
of instances that do not contain both instances. 
The following theorem states the relationship between neighbouring feature predictors in the original $d$-dimensional
space and the projected $d'$-dimensional space.

\begin{theorem}
\label{th:LSH}
Consider two (possibly nonlinear) feature predictors $h_{i}(\vec{x}) = \sigma(\vec{\mu}_{i}\T\vec{x})$,
and  $h_{j}(\vec{x}) = \sigma(\vec{\mu}_{j}\T\vec{x})$, parametrized by $\vec{\mu}_{i}, \vec{\mu}_{j} \in \R^{d}$,
and a transformation function $\sigma(\cdot) \in \{1,0\}$.
Let $\theta(\vec{\mu}_{i}, \vec{\mu}_{j})$ be the angle between $\vec{\mu}_{i}$ and $\vec{\mu}_{j}$.
The following relation holds between $\theta(\vec{\mu}_{i}, \vec{\mu}_{j})$ 
and the probability of agreement $p\left( \vec{h}_{i}(\cD_{(i,j)}) = \vec{h}_{j}(\cD_{(i,j)}) \right)$,
\[
	\theta(\vec{\mu}_{i}, \vec{\mu}_{j}) = \pi \left(1 - {p\left( \vec{h}_{i}(\cD_{(i,j)}) = \vec{h}_{j}(\cD_{(i,j)}) \right)}^{1/d'}\right) .
\]
\end{theorem}

The proof of Theorem~\ref{th:LSH} is given below, and follows from the properties of
locality sensitive hashing (LSH)~\cite{He:NIPS:2003,Andoni:CACM:2008,Indyk:STOC:98}.

\subsection*{Proof of Theorem~1}

Let us consider the agreement of the feature predictors $h_{i}$ and $h_{j}$ on the
$k$-th instance $\vec{x}_{k} \in \cD_{(i,j)}$. The probability of agreement can be written as,
\begin{equation}
\label{eq:agreement}
p\left( h_{i}(\vec{x}_{k}) = h_{j}(\vec{x}_{k}) \right) = 1 - p\left( h_{i}(\vec{x}_{k}) \neq h_{j}(\vec{x}_{k}) \right) .
\end{equation}
From the symmetry in the half-plane, the disagreement probability on the right side in \eqref{eq:agreement} can be written as
twice the probability of one parameter vector being projected positive and the other negative, given by:
\begin{equation}
\label{eq:double}
 p\left( h_{i}(\vec{x}_{k}) \neq h_{j}(\vec{x}_{k}) \right) = 2 p\left( \vec{\mu}_{i}\T\vec{x}_{k} \geq 0, \vec{\mu}_{j}\T\vec{x}_{k} < 0 \right) 
 \end{equation}
However, the vector $\vec{x}_{k}$ must exist inside the dyhedral angle $\theta(\vec{\mu}_{i}, \vec{\mu}_{j})$
formed by the intersection of the two half-panes spanned by $\vec{\mu}_{i}$ and $\vec{\mu}_{j}$. 
Therefore, the probability in \eqref{eq:double} can be estimated as the ratio between angles given by,
\begin{equation}
\label{eq:angle}
 p\left( \vec{\mu}_{i}\T\vec{x}_{k} \geq 0, \vec{\mu}_{j}\T\vec{x}_{k} < 0 \right)  = \frac{\theta(\vec{\mu}_{i}, \vec{\mu}_{j})}{2\pi} .
 \end{equation}
 From \eqref{eq:agreement}, \eqref{eq:double}, and \eqref{eq:angle}, we obtain,
 \begin{equation}
 \label{eq:full}
 p\left( h_{i}(\vec{x}_{k}) = h_{j}(\vec{x}_{k}) \right)  = 1 - \frac{\theta(\vec{\mu}_{i}, \vec{\mu}_{j})}{\pi} .
\end{equation}
If we assume that the instances in $\cD_{(i,j)}$ are i.i.d., then the agreement of the entire two $d'$-dimensional label vectors
can be computed as the product of agreement probabilities of each dimension, given by,
\begin{eqnarray}
\label{eq:prod}
p\left( \vec{h}_{i}(\cD_{(i,j)}) = \vec{h}_{j}(\cD_{(i,j)}) \right) &=& \prod_{k=1}^{d'} p\left( \vec{h}_{i}(\vec{x}_{k}) = \vec{h}_{j}(\vec{x}_{k}) \right) \nonumber \\
&=& {\left( 1 - \frac{\theta(\vec{\mu}_{i}, \vec{\mu}_{j})}{\pi} \right)}^{d'} .
\end{eqnarray}
From \eqref{eq:prod} it follows that,
\[ 
\theta(\vec{\mu}_{i}, \vec{\mu}_{j}) = \pi \left(1 - {p\left( \vec{h}_{i}(\cD_{(i,j)}) = \vec{h}_{j}(\cD_{(i,j)}) \right)}^{1/d'} \right)
\qed
\]

Theorem~\ref{th:LSH} states that we can measure the agreement between labels predicted by two feature predictors
using the angle between their corresponding parameter vectors.
More importantly, Theorem~\ref{th:LSH} provides us with a heuristic to approximately find the nearest neighbours of
each vertex without having to compute the confusion matrices for all pairs of vertices in the ClassiNet.
We compute the nearest neighbours for each feature predictor in the $d'$-dimensional space. 
Computation of ${p\left( \vec{h}_{i}(\cD_{(i,j)}) = \vec{h}_{j}(\cD_{(i,j)}) \right)}$ is closely related to the
calculation of hamming distance between the label vectors $\vec{h}_{i}(\cD_{(i,j)})$ and $ \vec{h}_{j}(\cD_{(i,j)})$.
The Point Location in Equal Balls (PLEB) algorithm~\cite{Indyk:STOC:98}
 can be used to compute the hamming distance in an efficient manner.
This algorithm considers random permutations of the bit streams and their sorting to find the vector with the 
closest hamming distance~\cite{Charikar:STOC:2002}. 
We use the variant of this algorithm proposed by Ravichandran and Hovy~\cite{Ravichandran:ACL:2005}
that extends the original algorithm to find the $k$-nearest neighbours.
Specifically, we use this algorithm to find the $k$-nearest neighbours for each feature $v_i$, and
compute edge-weights $w_{ij}$ for each $v_i$ and its nearest neighbours $v_j$ using the contingency table. 
Note that although we find the nearest neighbours using the approximate method described above, the edge-weights
computed between the selected neighbours are precise because they are based on the confusion matrix.

To estimate the size of the neighbourhood $k$ that we must select in order to obtain a reliable approximation of the
neighbours that we would have in the original $d$-dimensional space, we use the following procedure.
First, we randomly select  a small number $\alpha (\ll N)$ of vertices from the trained ClassiNet, and compute the confusion matrices
with each of those $\alpha$ vertices and the remainder of the vertices in the ClassiNet. 
We then compute the weights $w_{ij}$ of the edges that connect the selected $\alpha$ vertices to the rest of the vertices in the ClassiNet.
Following this procedure we compute the nearest neighbours of each vertex in $\alpha$
 without using the projection trick described above.
Second, we apply the projection method described above for all the vertices in the ClassiNet, and compute the nearest neighbours of
the $\alpha$ vertices that we selected. We then compare the overlap between the two sets of neighbourhoods.
In our preliminary experiments, we found that setting 
the neighbourhood size $k = 10$ to be an admissible trade-off between the accuracy of the neighbourhood computation and
the speed. Therefore, all experiments described in the paper use edge-weights computed with this $k$ value.

\subsection{ClassiNets vs. Co-occurrence Graphs}
\label{sec:classi-cooc}

Before we describe how to use the trained ClassiNets to classify short-texts, it is worth discussing 
the connection between word co-occurrence graphs and ClassiNets. 
Representing the association between words using co-occurrence graphs has a long history in NLP~\cite{Rada:2011}.
Word co-occurrences could be measured using symmetric measures, such as the Pointwise Mutual Information (PMI),
Log-Likelihood Ratio (LLR), or asymmetric measures such as KL-divergence, or conditional probability~\cite{FSNLP}.
In a co-occurrence graph, vertices correspond to words, and the weight of the edge connecting two vertices
represents the strength of association between the corresponding two words.
However, in a co-occurrence graph, two words $v_i$ and $v_j$ to be connected by an edge,
$v_i$ and $v_j$ must explicitly co-occur within the same context.

On the other hand, in ClassiNets, we have edges between vertices not only for the words that co-occur within the same
context, but also if they are predicted for the same instance even though
none of those features might actually be occurring in that instance. For example, for an instance $\vec{x}$ where
$x_i = x_j = 0$, we might still have $h_i(\vec{x}) = h_j(\vec{x}) = 1$. 
Therefore, ClassiNets consider implicit occurrences of features which would not be captured by co-occurrence graphs.
In fact, ClassiNets can be thought to be a generalized version of co-occurrence graphs that subsumes explicit co-occurrences.
To see this, let us define feature predictors $h_i$ and $h_j$ as follows:
\begin{eqnarray}
h_i(\vec{x}) = \vec{1}[x_i \neq 0] \\
h_j(\vec{x}) = \vec{1}[x_j \neq 0]
\end{eqnarray}
Here, $\vec{1}$ is the indicator function defined as follows:
\begin{equation}
\label{eq:indicator}
\vec{1}(\delta) = \begin{cases}
	1 & \delta = \text{TRUE} \\
	0 & \delta = \text{FALSE}
\end{cases}
\end{equation}

Then, $M_{11}$ in Table~\ref{tbl:conf} can be written as,
\begin{equation}
M_{11} = \sum_{\vec{x} \in \cD_{(i,j)}} \vec{1}[x_i \neq 0] \vec{1}[x_j \neq 0] ,
\end{equation}
which is the number of instances in which both features $v_i$ and $v_j$ would co-occur.
Therefore, ClassiNet reduces to co-occurrence graphs when the feature predictor is simply the indicator function for a single
feature. However, in general, feature predictors would consider not just a single feature but a combination (potentially non-linear)
of multiple features, thereby capturing broader information than in a word co-occurrence graph.

\section{Feature Expansion}
\label{sec:classinet:expand}

In this Section, we describe several methods to use the ClassiNets created in Section~\ref{sec:classinets}
for predicting missing features in instances, thereby overcoming the feature sparseness problem.
We refer to this operation as \emph{feature expansion}.
Given a train or a test instance $\vec{x} = (x_1, \ldots, x_d)\T$, we use the non-zero features, $x_i \neq 0$ in $x$ and find
similar vertices $v_j \in \cV$ from the created ClassiNet. In Section~\ref{sec:local}, we describe \emph{local feature expansion}
methods that consider only the nearest neighbours of the vertices in the ClassiNet that correspond to non-zero features in 
an instance, whereas in Section~\ref{sec:global} we propose a \emph{global feature expansion} method that propagates the original features
across the ClassiNet to predict the related features.

\subsection{Local Feature Expansion}
\label{sec:local}

Given a ClassiNet, we propose several feature expansion methods that consider the local neighbourhood of the non-zero features
that occur in an instance. We refer to such methods collectively as \emph{local feature expansion} methods.

\subsubsection{Independent Expansion}
\label{sec:expand:independent}
The first local feature expansion method we propose expands each feature in an instance independently of the others.
Specifically, we predict whether $v_i$ occurs in a given instance $\vec{x}$ using the feature predictor $h_i$ we 
trained from the unlabeled instances. If $h_i(\vec{x}) = 1$, then we append $v_i$ as an expansion feature to $\vec{x}$,
otherwise we ignore $v_i$. We repeat this process for all the vertices $v_i \in \cV$ and append the positively predicted vertices
to the original instance $\vec{x}$. 
If the $i$-th feature $x_i$ already appears in $\vec{x}$ and also predicted by $h_i(\vec{x})$
then we set its feature value to $x_i + h_i(\vec{x})$.
In the case where we have binary feature representations we will have $x_i \in \{0,1\}$.
Therefore, in the binary feature setting if a feature that already exists in an instance is predicted, then it will result in
doubling the feature weight ($\because x_i + h_i(\vec{x}) = 1 + 1 = 2$).
Moreover, instead of predicting the label, in a probabilistic classifier, such as the logistic regression, we can use
the posterior probability instead of the predicted label as $h_i(\vec{x})$ to compute feature values for the expansion features.

\subsubsection{Local Path Expansion}
\label{sec:expand:local}

This method extends the independent expansion method described in Section~\ref{sec:expand:independent} by
including all the vertices along the shortest paths that connect predicted features to the original features over the ClassiNet.
For example, let us assume that a feature $x_i = 0$ in an instance $\vec{x}$. If $h_i(\vec{x}) = 1$, we will append 
$v_i$ as well as all the vertices along the shortest paths that connect
$v_i$ to each feature $x_j \neq 0$ that exists in the instance $\vec{x}$.
Because all expanded features are connected to the original non-zero features that exist in the instance via some local path,
we refer to this approach as the \emph{local path expansion}. By construction, the set of expansion candidates produced by the
local path expansion method subsumes that of the independent expansion method.

\subsubsection{All Neighbour Expansion}
\label{sec:expand:nn}

In this expansion method, first, we use edge-weights to find the $k$-nearest neighbours of each vertex $v_i$, 
and connect all the neighbours for each vertex to create a $k$-nearest neighbour graph 
from the trained ClassiNet. The $k$-nearest neighbour graph that we create from the ClassiNet
in this manner is a subgraph of the ClassiNet. Two vertices $v_i$ and $v_j$ are connected by an edge in this $k$-nearest
neighbour graph if and only if $v_i$ is among the top $k$ most similar vertices to $v_j$ as well as
$v_j$ is among the top $k$ most similar vertices to $v_i$.
The weights of all the edges in this $k$-nearest neighbour graph are set to $1$.

Next, for each non-zero feature in an instance $\vec{x}$, we use its nearest neighbours
as expansion features. This method ignores the absolute values of the edge-weights in the ClassiNet, and considers only their
relative strengths. If we increase the value of $k$, we will have a larger set of candidate expansion features.
However, it will also result in considering less relevant features to the original features.
Therefore, there exists a trade-off between the number of expansion candidates we can use for feature vector expansion,
and the relevancy of the expansion features to the original features. Using development data, we constructed $k$-nearest
neighbour graphs for varying $k$ values, and found that $k > 4$ settings often result in noisy neighbourhoods.
Consequently, when using neighbour expansion, we set $k = 4$.

\subsubsection{Mutual Neighbour Expansion}
\label{sec:expand:mutual}

The mutual neighbour expansion method also uses the same $k$-nearest neighbour graph as used by the 
all neighbour expansion method described in Section~\ref{sec:expand:nn}.
The mutual neighbour expansion method selects a vertex $v_j$ in ClassiNet as an expansion candidate, if there exists at least two distinct
vertices $v_i$,  $v_k$ in the ClassiNet for which $x_i \neq 0$, and $x_k \neq 0$ in the instance $\vec{x}$ to be expanded. 
This method can be seen as a conservative version
of the all neighbour expansion method described in Section~\ref{sec:expand:nn} because, we would ignore vertices $v_j$ that are
nearest neighbours of only a single feature in the original feature vector. The mutual neighbour expansion method addresses 
the issue associated with previously proposed local feature expansion methods,
which select expansion candidates
separately for each non-zero feature in the feature vector to be expanded, ignoring the fact that the feature vector represents
a single coherent short-text. However, this conservative expansion candidate selection strategy of the
 mutual neighbour expansion method means that we will have a smaller set of expansion candidates in comparison to,
 for example, the all neighbour expansion method. 

\subsection{Global Feature Expansion}
\label{sec:global}

The local feature expansion methods described in Section~\ref{sec:local} consider only the vertices in the ClassiNet that 
are \emph{directly connected} to a feature in an instance as expansion candidates.
Even in the case of local path expansion (Section~\ref{sec:expand:local}), the expansion candidates are limited to the
local neighbours of the original features and the predicted features.
Considering that ClassiNet is a directed graph, we can perform label propagation on ClassiNet to find features that are
not directly connected nor appearing in the local neighbourhood of a feature in a short-text but still relevant.

For example, assume that \emph{Google} and \emph{Microsoft} are not local neighbours in a ClassiNet.
Consequently none of the local neighbour expansion methods will be able to predict \emph{Microsoft} as a relevant feature for
expanding a short-text containing \emph{Google}. However, if \emph{Bing}, a Web search engine similar to \emph{Google}, appears in the
local neighbourhood of \emph{Google} in the ClassiNet, and if we can propagate from \emph{Bing} to its parent company
\emph{Microsoft} via the ClassiNet, then we will be able to predict \emph{Microsoft} as a relevant feature for \emph{Google}.
The propagation might be over multiple hops, thereby reaching beyond the local neighbourhood of a feature.

Propagation over ClassiNet can also help to reduce the ambiguity in feature expansion.
For example, consider the sentence ``\emph{Microsoft and Apple are competing for the tablet computer market.}''.
If we do not perform word sense disambiguation prior to feature expansion, and we expand each feature independently of the others,
then it is likely that we might incorrectly expand \emph{apple} by other types of fruits such as \emph{banana} or \emph{orange}.
Such phenomena are observed in prior work on set expansion and is referred to as \emph{semantic drift}~\cite{Kozareva:NAACL:2010}.
However, if we find the expansion candidates jointly, such that they are relevant to all the features (words) in the sentence,
then they must be relevant to both \emph{Microsoft} as well as \emph{Apple}, which encourages other IT companies, such as
\emph{Google} or \emph{Yahoo} for example. All local feature expansion methods described in Section~\ref{sec:local} except the
independent expansion method address this issue by ranking expansion candidates depending on how well they are related to all the
features in a short-text. Label propagation can solve this ambiguity problem in a more systematic manner by converging multiple random walks
initiated at different features that exist in a short text. Next, we describe a \emph{global feature expansion} method based on
propagation over ClassiNet.

\begin{figure}[t]
\centering
\includegraphics[height=50mm]{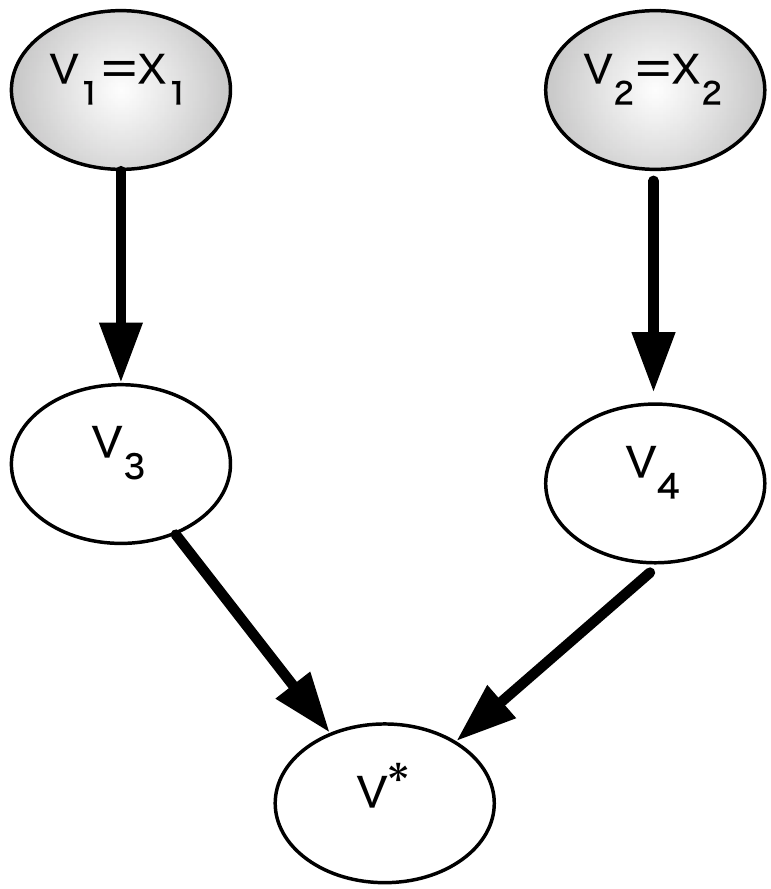}
\caption{Computing the feature value of an expansion feature $v^*$ for an instance that has $v_1 = x_1$ and
$v_2 = x_2$ as non-zero features.}
\label{fig:global}
\end{figure}

First, let us describe the proposed global feature expansion method using the ClassiNet shown in Figure~\ref{fig:global}.
Here, we consider expanding an instance $\vec{x} = (x_1, x_2)\T$ with two non-zero features $v_1 = x_1$ and $v_2 = x_2$
($x_1 \neq 0$, and $x_2 \neq 0$).
We would like to compute the likelihood $p(v^*|\vec{x})$ of a vertex $v^*$ as an expansion candidate for the instance $\vec{x}$.
From Figure~\ref{fig:global} we see that there are two possible paths reaching $v^*$ starting from the original features $x_1$ and $x_2$.
Assuming that the two paths are independent, we compute $p(v^*|\vec{x})$ as follows:
\begin{equation}
%\small
p(v^*|\vec{x}) = p(x_1)p(v_3|x_1)p(v^*|v_3) + p(x_2)p(v_4|x_2)p(v^*|v_4)
\label{eq:example}
\end{equation}

The computation described in Figure~\ref{fig:global} can be generalized for an arbitrary
 ClassiNet $\cG(\cV, \cE, \mat{W})$, and an instance $\vec{x} = (x_1, \ldots, x_d)\T$.
For this purpose, let us define the set of non-cyclic paths connecting two vertices $v_i$, $v_j$ in $\cG$ to be $\Gamma(v_i, v_j)$.
For the example shown in Figure~\ref{fig:global} we have the two paths $x_1 \rightarrow v_3 \rightarrow v^*$, and
$x_2 \rightarrow v_4 \rightarrow v^*$.
We compute the likelihood $p(v^* | \vec{x})$ of a vertex $v^* \in \cV$ being an expansion candidate of $\vec{x}$ as follows:
\begin{equation}
p(v^*|\vec{x}) = \sum_{k=1}^{d} \left( x_k p(x_k=v_k)\prod_{(a,b) \in \Gamma(x_k, v^*)}p(b|a) \right)
\label{eq:global}
\end{equation}
If a feature $x_k = 0$, then the likelihoods corresponding to paths starting from $x_k$ will be
 ignored in the computation of \eqref{eq:global}.
The prior probabilities of features $p(x_k)$ can be estimated from train data by dividing the 
number of instances that contain $x_k$ by the total number of instances. Alternatively, we could set a uniform prior
for $p(x_k)$ thereby considering all the words that occur in an instance equally. We follow the latter approach in our experiments.

The sum-product computation over paths can be efficiently computed by observing that it can be modeled
as a  label propagation problem over a directed weighted graph, where an instance $\vec{x}$ is the initial state vector
and the transition probabilities are given by the weight matrix $\mat{W}$. 
Vertices that can be reached after $q$ hops are given by $\sum_{i=1}^{q}\mat{W}^{i}\vec{x}$. 
Neighbours that are distantly located in the ClassiNet are less reliable as expansion candidates.
To reduce the noise due to distant (and potentially irrelevant) vertices during the propagation, we introduce a damping
factor $0 < \gamma \leq 1$ in the summation, $\sum_{i=1}^{q}\gamma^i \mat{W}^{i} \vec{x}$.
In Section~\ref{sec:damp}, we experimentally study the effect of the level of damping on the classification 
accuracy of short-text classification.

The feature expansion methods we described above are used to predict missing features for both train and test instances.
We expand feature vectors representing the train/test instances, and assign unique identifiers to the 
expansion features, thereby distinguishing between the original features and the expanded features.
For example, given the positive sentiment labeled train sentence ``\emph{I love dogs}'', we can represent it using the feature
vector, [(\emph{I}, 1), (\emph{love}, 1), (\emph{dog}, 1)]. 
Here, we assume that lemmatization has been conducted on the input and the feature \emph{dogs} has been converted to 
its singular form \emph{dog}. Let us further assume that from the trained ClassiNet we were able to predict that
\emph{cat} is a related feature for \emph{dog}, and the candidate score $p(cat|dog) = 0.8$.
Next, we add the feature (\emph{EXP=cat}, 0.8) to the feature vector representing this train instance,
where the prefix \emph{EXP=} indicates that it is a feature introduced by the expansion method and not a feature that existed in
the original train instance. Distinguishing original vs. expansion features is useful when we would like to learn different weights 
for the same feature depending on whether it is expanded or not.
For example, if a particular feature is not very useful as an expansion feature, it will be assigned a lower weight
thereby effectively pruning that feature out from the model learnt by the classifier.

% A note about what happens if a feature already exist in the original vector and is also predicted? We add the scores.
% We could do better if the classifier returns posterior probabilities. 
% We could down-weight but we do not do this because it adds additional parameters to tune.

The first step of learning a ClassiNet is learning the feature predictors. In this regard, any word embedding learning method can be used for the purpose of learning feature predictors. Once the feature predictors are learnt, we can create a ClassiNet in the same manner as we propose in this paper and use the ClassiNet created to perform feature expansion using local/global feature expansion methods we propose in the paper. This view of ClassiNets illustrates the general applicability of the proposed method. 

\section{A Theoretical Analysis of ClassiNets}
\label{sec:theory}

Before we empirically evaluate the performance of the proposed ClassiNets for feature expansion in short-text classification,
let us analyze some interesting properties of ClassiNets.
To simplify the analysis, let us assume that we are using a ClassiNet for learning a linear classifier $\vec{\phi} \in \R^{d}$ for a binary classification task.
Specifically, let us assume that we are given a train dataset
$\{(\vec{x}^{(k)}, y^{(k)})\}_{k=1}^{N}$ consisting of $N$ instances, where each train instance $k$ is represented by
a feature vector $\vec{x}^{(k)} \in \R^{d}$. The binary target label assigned to the $k$-th train instance is denoted by $y^{(k)} \in \{1, -1\}$.
For correctly classified train instances $\vec{x}^{(k)}$ we have, $y^{(k)}\phi\T\vec{x}^{(k)} > 0$.

We use the trained linear classifier $\vec{\phi}$, and predict the label $\hat{y}$ of an unseen test instance $\hat{\vec{x}}$ as follows:
\begin{eqnarray}
 \label{eq:pred}
 \hat{y} = \begin{cases} 
 	1 & \text{if } \phi\T\hat{\vec{x}} > 0 \\
	 -1 & \text{otherwise}
\end{cases}
\end{eqnarray}

Let us assume that we have learnt a feature predictor $h_{i}$ that predicts whether the $i$-th feature exists in a given instance.
As described in Section~\ref{sec:overview}, we can use any classification algorithm to learn the feature predictors.
However, as a concrete case, let us consider linear classifiers in this analysis. In the case of linear classifiers, we can represent the
feature predictor learnt for the $i$-th feature by the vector $\vec{\mu}_{i}$. Following the notation introduced in Section~\ref{sec:overview},
we can write the feature predictor $h_{i}$ as follows:
\begin{equation}
 h_{i} (\vec{x}) = \begin{cases} 1 & \text{if } \vec{\mu}_{i}\T\vec{x} > 0 \\
 -1 & \text{otherwise}
 \end{cases}
\end{equation}
In the ClassiNets described in the paper so far, we used the predicted discrete labels as the values of the predicted features during feature expansion.
However, in this analysis let us consider the more general case where we use the actual prediction score, $\vec{\mu}_{i}\T\vec{x}$ as the
contribution of the feature expansion towards the $i$-th feature.

We can construct the expanded feature vector, $\vec{x}^{*} \in \R^{d}$, of the feature vector $\vec{x} \in \R^{d}$ considering the
inner-product between $\vec{x}$ and each of the feature predictors $\vec{\mu}_{i}$ as in \eqref{eq:expand}.
\begin{equation}
\label{eq:expand}
 \vec{x}^{*} = [ (x_{1} + \vec{\mu}_{i}\T\vec{x}), \ldots, (x_{i} + \vec{\mu}_{i}\T\vec{x}), \ldots, (x_{d} + \vec{\mu}_{d}\T\vec{x})]\T
\end{equation}
Here, we denote the $i$-th dimension of the feature vector $\vec{x}$ by $x_{i}$.
We can transform the given train dataset $\{(\vec{x}^{(k)}, y^{(k)})\}_{k=1}^{N}$ by expanding each feature vector separately
using \eqref{eq:expand}, and use the expanded feature vectors to train a binary linear classifier $\vec{\phi}^{*}$.
Following \eqref{eq:pred}, we can use $\vec{\phi}^{*}$ to predict the label for a test instance $\vec{x}^{*}$ based on the prediction score
given by
\begin{eqnarray}
\vec{\phi}^{*}\T\vec{x}^{*} &=& \sum_{i=1}^{d} \phi_{i}^{*} \left( x_{i} + \vec{\mu}_{i}\T\vec{x} \right) \nonumber \\
&=& \sum_{i=1}^{d} \phi_{i}^{*} x_{i} + \sum_{i=1}^{d} \phi_{i}^{*}  \vec{\mu}_{i}\T\vec{x} \nonumber \\
&=& \vec{\phi}^{*}\T \vec{x} + \vec{\phi}^{*}\T \mat{L} \vec{x} \label{eq:exp2} \\
&=& \vec{\phi}^{*}\T \left(\mat{I} + \mat{L} \right) \vec{x} \label{eq:exp3}
\end{eqnarray}
Here, $\mat{I} \in \R^{d \times d}$ is a unit matrix, and
$\mat{L} \in \R^{d \times d}$ is the matrix formed by arranging the feature predictors $\vec{\mu}_{i}$ in rows.
In other words, $\mat{L} = [\vec{\mu}_{1} \ldots \vec{\mu}_{d}]\T$.

The first term in \eqref{eq:exp2} corresponds to classifying the non-expanded (original) instance $\vec{x}$ using the classifier trained using
the expanded train dataset. The second term in \eqref{eq:exp2} represents the prediction score due to feature expansion.
From \eqref{eq:exp3} we see that performing feature expansion on a feature vector $\vec{x}$ is equivalent to multiplying
the matrix $\left(\mat{I} + \mat{L} \right)$ into $\vec{x}$. Therefore, local feature expansion methods described in Section~\ref{sec:local}
can be seen as projecting the train feature vectors into the same $d$-dimensional feature space spanned by the features that
exist in the train instances. As a special case, we see that  when we do not learn feature predictors 
we have $\mat{L} = \mat{0}$, for which \eqref{eq:exp2} reduces to the prediction score 
$\vec{\phi}^{*}\T\vec{x}$ of the binary linear classifier trained using non-expanded train instances.

\subsection{Edge weights of ClassiNets}

Recall that, $w_{ij}$ the weight of the edge connecting the vertex $i$ to vertex $j$ in a ClassiNet was defined by \eqref{eq:weight}.
In the case of binary linear feature predictors $\vec{\mu}_{i}$ and $\vec{\mu}_{j}$ we considered in the previous section,
let us estimate the value of $w_{ij}$. Using the indicator function $\vec{1}$ defined by \eqref{eq:indicator}, we compute
$M_{11}$ and $(M_{11} + M_{10})$ in \eqref{eq:weight} as follows:
{\small
\begin{eqnarray}
 && M_{11} = \sum_{k=1}^{N} \vec{1}[(y^{(k)}\vec{x}^{(k)}\T\vec{\mu}_{i}\mkern-5mu>\mkern-5mu0) \land (y^{(k)}\vec{x}^{(k)}\T\vec{\mu}_{j} \mkern-5mu > \mkern-5mu 0)] \label{eq:M11} \\
 && M_{11} + M_{10} =  \sum_{k=1}^{N} \vec{1}[(y^{(k)}\vec{x}^{(k)}\T\vec{\mu}_{i} > 0)] \label{eq:M*}
\end{eqnarray}
}
Let us assume that we sample instances $\vec{x}$ from the train dataset randomly according to the distribution $p(\vec{x})$.
Then the expected counts in $\hat{M}_{11}$ and $\hat{M}_{10}$ in  \eqref{eq:M11} and \eqref{eq:M*} can be expressed using the
expected number of the correct classifications made by the feature predictors $\vec{\mu}_{i}$ and $\vec{\mu}_{j}$ as follows:
{\small
\begin{eqnarray}
&& \hat{M}_{11} = \Ep_{p(\vec{x})}\left[ \vec{1}[(y\vec{x}\T\vec{\mu}_{i} > 0) \land (y\vec{x}\T\vec{\mu}_{j} > 0)] \right] \label{eq:M11:hat} \\
&& \hat{M}_{11} + \hat{M}_{10} =  \Ep_{p(\vec{x})} \left[ \vec{1}[(y\vec{x}\T\vec{\mu}_{i} > 0)] \right] \label{eq:M*:hat}
\end{eqnarray}
}
Using the expected counts given by \eqref{eq:M11:hat} and \eqref{eq:M*:hat} we can compute the approximate value of
the edge weight $\hat{w}_{ij}$ as follows:
\begin{equation}
 \label{eq:weight:approx}
 \hat{w}_{ij} = \frac{\Ep_{p(\vec{x})}\left[ \vec{1}[(y\vec{x}\T\vec{\mu}_{i} > 0) \land (y\vec{x}\T\vec{\mu}_{j} > 0)] \right]} { \Ep_{p(\vec{x})} \left[ \vec{1}[(y\vec{x}\T\vec{\mu}_{i} > 0)] \right]}
\end{equation}

If we have a sufficiently large train dataset, then \eqref{eq:weight:approx} provides an alternative procedure for estimating the edge weights.
We could randomly select samples from the train dataset, predict the features $i$ and $j$ for those samples, and compute the expectations as
ratio counts. We can repeat this procedure many times to obtain better approximations for the edge weights.
Although this is a theoretically feasible procedure for approximately computing the edge weights, it can be slow in practice and
might require many samples before we obtain a reliable approximation for the edge weights.
Therefore, the edge weight computation method described in Section~\ref{sec:project} is more appropriate for practical purposes.

\subsection{Analysis of the Global Feature Expansion Method}

We already showed in \eqref{eq:exp3} that local feature expansion methods can be considered as feature vector transformation
methods by a matrix $(\mat{I} + \mat{L})$. However, an important strength of ClassiNet is that we can  propagate the
predicted features over the network using the global feature expansion method described in Section~\ref{sec:global}.

Let us denote the edge-weight matrix of the ClassiNet $\cG$ by $\mat{W}$. The $(i,j)$-th element of $\mat{W}$ is denoted by
$w_{ij}$. The connection between edge weights $w_{ij}$ and the feature predictors $\vec{\mu}_{i}$ and $\vec{\mu}_{j}$ is given
by \eqref{eq:weight:approx}. In the global feature expansion method, we repeatedly propagate the predicted features across the
network, which can be seen as a repeated multiplication using $\gamma \mat{W}$, where $\gamma$ is the damping factor
described in Section~\ref{sec:global}. Observing this connection, we can derive the prediction score under the global feature expansion
method similar to \eqref{eq:exp3} as follows:
\begin{eqnarray}
 \vec{\phi}^{*}\T\vec{x}^{*}&=& \vec{\phi}^{*}\T \left(\mat{I} + \gamma\mat{W} + \ldots + \gamma^{q} \mat{W}^{q} \right) \vec{x} \nonumber \\
 &=&  \vec{\phi}^{*}\T (\mat{I} - \gamma \mat{W})\inv (\mat{I} - \gamma^{(q+1)} \mat{W}^{(q+1)}) \vec{x} \label{eq:exp4}
\end{eqnarray}
For the summation shown in \eqref{eq:exp4} to hold, and the matrix $(\mat{I} - \gamma \mat{W})$ to be invertible,
for all eigenvalues $\lambda_{r}$ of $\mat{W}$ we require $\gamma |\lambda_{r}| < 1$.
This requirement can be met in practice by a sufficiently small damping factor. For example, we could set $\gamma = 1/(1 + |\lambda_{\max}||)$,
where $|\lambda_{\max}|$ is the eigenvalue of $\mat{W}$ with the maximum absolute value.

As a special case where we propagate the features without truncating, we have $q \rightarrow \infty$, for which we obtain the
prediction score given in \eqref{eq:inf}.
\begin{equation}
\label{eq:inf}
  \vec{\phi}^{*}\T\vec{x}^{*} =  \vec{\phi}^{*}\T (\mat{I} - \gamma \mat{W})\inv  \vec{x}
\end{equation}
From \eqref{eq:inf}, we see that, similar to the local feature expansion methods, the global feature expansion method can also be seen as
projecting the input feature vector $\vec{x}$ using the matrix $(\mat{I} - \gamma \mat{W})\inv$.

\section{Experiments}
\label{sec:exp}
% Using SCL is appropriate because (a) it has been used previously on cross-domain sentiment classification 
% where the feature sparseness is an issue and (b) in the context of binary sentiment classification, similar to our evaluation task.

We create a ClassiNet using 257,306 unlabeled sentences from the
 Large Movie Review dataset\footnote{\url{http://ai.stanford.edu/~amaas/data/sentiment/}}.
%We selected the most frequent 1000 unigrams in the dataset and created a ClassiNet that has $1000$ vertices and $489,300$ edges. 
Each word in this dataset is uniquely represented by a vertex in the ClassiNet. We learn linear predictor for each feature
using automatically selected positive (reviews where the target feature appears) and negative (reviews where the target feature does not appear)
training instances.  The ClassiNet created from this dataset contains $489,000$ vertices.
This ClassiNet is used in all the experiments described in the remainder of this paper.

For evaluation purposes we use four binary classification datasets:
the Stanford sentiment treebank (\textbf{TR})\footnote{\url{http://nlp.stanford.edu/sentiment/treebank.html}} (903 positive test instances and
903 negative test instances),
movie reviews dataset (\textbf{MR})~\cite{Pang:ACL:2005} (5331 positive instances and 5331 negative instances), 
customer reviews dataset  (\textbf{CR})~\cite{Hu:KDD:2004} (925 positive instances and 569 negative instances), and
subjectivity dataset (\textbf{SUBJ})~\cite{Pang+Lee:04a} (5000 positive instances and 5000 negative instances).
We perform five-fold cross-validation in all datasets, except in the Stanford sentiment treebank 
where there exists a pre-defined test and train split.
In each dataset, we use the train portion to learn a binary classifier. Next, we use the trained ClassiNet to expand the
feature vectors for the test instances. We then measure the classification accuracy of the binary classifier on the
expanded test instances. If high classification accuracies are obtained using a particular feature expansion method,
then that feature expansion method is considered superior.

We use a CPU server containing 48 cores of 2.5GHz Intel Xeon CPU and 512GB RAM in our experiments.
The entire training pipeline of training feature predictors, building the ClassiNet and expanding training instances using Global feature expansion method takes approximately 1.5 hours. The testing phase is significantly faster because we can use the created ClassiNet to expand test instances
and use the trained model to make predictions. For example, for the \textbf{SUBJ} dataset, which is the largest among all datasets used in our experiments, it takes only 5 minutes to both expand (using Global feature expansion) and predict (using logistic regression).

%%%%%%%%%%%%%%%% dataset statistics %%%%%%%%%%%%%%%%%%%%%%%%%%%%%%%
% MR pos = 5331, neg = 5331, train = 8530, test = 2132
%
% CR pos = 925, neg = 569. train = 1196, test = 298  
% 
% SUBJ subjective (pos) = 5000, objective (neg) = 5000, train = 8000, test = 2000
%
% we selected sentences with polarity annotations. All positive polarities were grouped to create the
% positive class and all negative polarities were grouped to create the negative class.
%%%%%%%%%%%%%%%%%%%%%%%%%%%%%%%%%%%%%%%%%%%%%%%%%%%%%%%%%%

\subsection{Binary Classification of Short-Texts}
\label{sec:sentiment}

Direct evaluation of the features predicted by the ClassiNet is difficult because there is no gold standard for feature expansion.
Instead, we perform an extrinsic evaluation of the created ClassiNet by using it to expand feature vectors representing sentences in 
several binary text classification tasks. If we can observe any increase (or decrease) in classification accuracy
for the target classification task when we use the features predicted by the ClassiNet, then it can be directly associated with
the effectiveness of the ClassiNet.
For the purpose of training a binary classifier, we represent a sentence by a real-valued vector, in which elements correspond to the
unigrams extracted from that sentence. The feature values are computed using the tfidf measure.
We train a binary logistic regression model, where the $L_{2}$ regularisation coefficient is tuned using development data selected from
the Stanford sentiment treebank dataset. 

We use classification accuracy, which is defined as the ratio between the correctly classified test
sentences and the total number of test sentences in the Stanford sentiment treebank. 
In addition to reporting the overall classification accuracies, we report classification accuracies
separately for the positively labeled instances and the negatively labeled sentences. 
%We use the Clopper-Pearson confidence intervals~\cite{Clopper-Pearson}
%computed at the $5\%$ confidence level to test for statistical significance.
Because this is a binary classification task, a random classifier would obtain an accuracy of $50\%$.
There are $903$ positive and $908$ negative sentiment labeled test sentences in the Stanford sentiment treebank
test dataset. Therefore, a baseline that assigns the majority label would obtain an accuracy of $50.13\%$ on this dataset.

Table~\ref{tbl:sentiment} compares the sentiment classification accuracies obtained by the following methods:

\textbf{No Expansion:} This baseline does not perform any feature expansions. It trains a binary logistic regression classifier
using the train sentences, and applies it to classify sentiment of the test sentences. This baseline demonstrates the level
of performance we would obtain if we had not performed any feature expansion. It can be seen as a lower-baseline for this task.

\textbf{Independent Expansion:} This method is described in Section~\ref{sec:expand:independent}. 

\textbf{Local Path Expansion:} This method is described in Section~\ref{sec:expand:local}.

\textbf{All neighbour Expansion:} This method is described in Section~\ref{sec:expand:nn}.

\textbf{Mutual neighbour Expansion:} This method is described in Section~\ref{sec:expand:mutual}.

\textbf{WordNet:}
Using lexical resources such as thesauri to find related words is a popular
 technique used in query expansion~\cite{Fang:ACL:2008,Gong:2005}.
To simulate the performance that we would obtain if we had used an external resource such as
the WordNet to find the expansion candidates, we implement the following baseline. 
In the WordNet, words that are semantically related are grouped into clusters called \emph{synsets}.
For each feature in a test instance, we search the WordNet for that feature, 
and use all words listed in synsets for that feature as its expansion candidates.
We consider all synonyms in a synset to be equally relevant as expansion candidates of a feature.

\textbf{SCL:}
Domain adaptation methods attempt to overcome the feature mismatch between source and target domains
by predicting missing features and/or learning a lower-dimensional embedding common to the two domains.
Although we do not have two domains in our setting, we can still apply domain adaptation methods such as the
structural correspondence learning (SCL) proposed by Blitzer et al.~\cite{Blitzer:EMNLP:2006} to predict missing features
in a given short-text. 
SCL was described in detail in Section~\ref{sec:related}.
Specifically, we train SCL using the same set of vertices as used by the ClassiNet as pivots.
This enables us to conduct a fair comparison between SCL and methods that use ClassiNet because the
performance between SCL and methods that use ClassiNet can be directly attributable to the
projection method used in SCL and not due to any differences of the expansion set.
We then train linear predictors for those pivots using logistic regression.
We arrange the trained linear predictors as rows in a matrix, on which we subsequently perform singular value decomposition
to obtain a lower-dimensional projection. Following the recommendations in  \cite{Blitzer:EMNLP:2006},
we set the dimensionality of the projection to $50$.
Both train and test instances are first projected to this lower-dimensional space
and we append the projected features to the original feature vectors.
Next, we train a binary sentiment classifier using logistic regression with $\ell_{2}$ regularisation. 
The regularisation coefficient is set using a held-out set of review sentences.

\textbf{FTS:}
FTS is the frequent term sets method proposed by Man~\cite{Man:2014}.
First, co-occurrence and class-orientation relations are defined among features (terms).
Next,  terms that are frequent in those relations more than a pre-defined threshold (support) are selected as expansion candidates.
Finally, for each feature in a short text, the frequent term sets containing this feature are appended as expansion 
features to the original feature vector representing the short-text.
FTS can be considered as a method that uses clusters of features induced from the data instances to overcome
the feature sparseness problem.

\textbf{CBOW:}
To compare the explicit feature expansion approach used by ClassiNets against implicit text representation methods,
we use pre-trained word embeddings to represent a short-text in a lower-dimensional space.
Specifically, we create $300$ dimensional word embeddings using the same corpus used by ClassiNets to create continuous bag-of-words (CBOW) ~\cite{Milkov:2013} embeddings, and add the word embedding vectors for all the words in a short text to
create a $300$ dimensional vector that represents the given short-text.

\textbf{Global Feature Expansion:}
This method propagates the original features across the trained ClassiNet, and is described in Section~\ref{sec:global}.
It is the main method proposed in this paper.

\begin{table}[t]
\caption{Binary classification accuracies.}
\begin{center}
\begin{tabular}{l c c c c} \toprule
Method							&	\textbf{TR}	& \textbf{MR} 	& \textbf{CR}	&	\textbf{SUBJ} \\ \midrule
No Expansion						&	$76.31$		& $73.35$		&  $81.54$	&	$88.95$			\\
Independent Expansion				&	$75.32$		&  $74.11$		&   $78.19$	&	$87.15$		\\	
Local Path Expansion				&	$76.97$		& $73.73$		&  $81.87$	&	$88.05$			\\
All neighbour Expansion			&	$77.36$		& $72.93$		&  $82.55$	&	$88.75$			\\
Mutual neighbour Expansion			&	$77.13$		& $74.15$		&  $80.87$	&	$88.95$			\\
WordNet							&	$76.58$ 		& $66.09$		&  $79.86$	&	$77.95$			\\
SCL~\cite{Blitzer:EMNLP:2006}		& 	$78.02$ 		& $74.44$		&  $81.20$	&	$89.25$			\\
FTS~\cite{Man:2014}				&	$76.47$ 		& $66.83$		&  $62.41$	&	$50.15$			\\
CBOW							&	$77.52$		& $73.31$		&  $79.87$	&	$88.88$			\\				
Global Feature Expansion			&	$\mathbf{78.30}$		& $\mathbf{81.20}^{*}$		&  $\mathbf{83.89}^{*}$	&	$\mathbf{89.70}$			\\ \bottomrule
\end{tabular}
\end{center}
\label{tbl:sentiment}
\end{table}

% Proposed is best in all four datasets, significant in MR and CR shown by *
% Among local expansion methods all neighbour expansion is the best in TR, CR, whereas mutual neighbour expansion is the best in MR, SUBJ
% Local expansion methods are unable to outperform no expansion baseline in SUBJ. --> Global expansion is vital
% CBOW and SCL are doing better than local expansion methods. This shows that implicit representations (embeddings) are 
% competitive than local feature expansion methods. However, the Global feature expansion is doing better than SCL and CBOW.
% It is not just sufficient to create a ClassiNet, but it is equally important to perform global expansions.
% FTS is worst. Simply expanding using frequent features is not adequate.
% WordNet uses manually created synsets and outperform FTS in all cases.
% Proposed method does not require manually created synsets and performs better than WordNet.

We summarise the classification accuracies obtained with different approaches discussed on the four test datasets in Table~\ref{tbl:sentiment}.
For each dataset we indicate the best performing method using boldface font, whereas an asterisk indicates if the best performance
reported is statistically significantly better than the second best method on the same dataset according to 
a two-tailed paired t-test under $0.01$ confidence level.
From Table~\ref{tbl:sentiment}, we see that the proposed \textbf{Global Feature Expansion} method obtains the best performance
in all four datasets. Moreover, in \textbf{MR} and \textbf{CR} datasets its performance is significantly better than the second best
methods (respectively \textbf{SCL} and \textbf{All Neigbour Expansion}) on those two datasets .

Among the four local expansion methods, \textbf{All neighbour Expansion} reports the best performance in \textbf{TR} and
\textbf{CR} datasets, whereas the \textbf{Mutual neighbour Expansion} reports the best performance in \textbf{MR} and \textbf{SUBJ} datasets.
\textbf{Independent Expansion} method performs worse than the \textbf{No Expansion} baseline in \textbf{TR}, \textbf{CR}, and \textbf{SUBJ}
datasets indicating that by individually expanding each feature in a short-text we introduce a significant level of noise into the short-text.
This result shows the importance for a feature expansion methods to consider all the features in an instance when adding related features to an instance.
None of the local feature expansion methods are able to outperform the global feature expansion method in any of
the datasets. In particular, in the \textbf{SUBJ} dataset we see that none of the local feature expansion methods outperform the 
\textbf{No Expansion} baseline. This result implies that it is not sufficient to simply create a ClassiNet, but it is also important
to use an appropriate feature expansion method on the built ClassiNet to find expansion features 
to overcome the feature sparseness problem in short-text classification.

\textbf{FTS} method performs poorly in all our experiments. This indicates that the frequency of a feature is not a good indicator of its
effectiveness as an expansion candidate. 
On the other hand, \textbf{WordNet} method that uses synsets as expansion candidates performs much better than \textbf{FTS} method.
Not surprisingly, this result shows that synonyms are useful as expansion candidates. 
However, a prerequisite of this approach is the availability of a thesauri that are either manually or semi-automatically created.
Such linguistic resources might not be available or incomplete for some languages.
On the other hand, our proposed method does not require such linguistic resources.

\textbf{CBOW} and \textbf{SCL} methods perform competitively with the \emph{Global Feature Expansion} method in all datasets.
Given that both \textbf{CBOW} and \textbf{SCL} are using word-level embeddings to compute a representation for a short text,
this result shows the effectiveness of word-level embeddings as a method to overcome feature sparseness in short-text classification tasks.
We compare non-compositional sentence-level embedding methods against the proposed \textbf{Global Feature Expansion} method later
in Section~\ref{sec:sentemb}.

\subsection{Comparisons against sentence-level embeddings}
\label{sec:sentemb}
% comparison against skip-thought, paragraph vect, and FastSent using published results.

An alternative direction for representing short-texts is to project the entire text directly to a lower-dimensional space,
without applying any compositional operators to word-level embeddings.
The expectation is that the overlap between short-texts in the projected space will be higher than that in the
original space such as a bag-of-word representation of a short-text.
Skip-thought vectors~\cite{Kiros:2015}, FastSent~\cite{Hill:NAACL:2016}, and Paragraph2Vec~\cite{Le:ICML:2014}
 are popular sentence-level embedding methods that have reported state-of-the-art performance on text classification tasks.
In contrast to our proposed method which explicitly append features to the original feature vectors to overcome the feature sparseness
problem, sentence-level embedding methods can be seen as an implicit feature representation method.

In Table~\ref{tbl:sentemb}, we compare the proposed method against the state-of-the-art sentence-level embedding methods.
We use the published results in \cite{Kiros:2015} on \textbf{MR}, \textbf{CR}, and \textbf{SUBJ} datasets for 
Skip-thought, FastSent, and Paragraph2Vec, without re-training those methods. All three methods are trained on the Toronto books corpus~\cite{moviebook}.
Performance of these methods on the \textbf{TR} dataset were not available. 
As a multiclass classification setting, we used the \textbf{TREC} question-type classification dataset.
In this dataset, each question is manually classified to 6 question types
depending on the information asked in the question such as abbreviation, entity, description, human, location and numeric.
We use the same classinet as we used in the binary classification tasks to predict features for 5500 train and 500 test questions.
A multiclass logistic regression classifier is trained on feature vectors with missing features predicted and tested on the feature vectors for the test questions with
missing features predicted.

Next, we briefly describe the methods compared in Table~\ref{tbl:sentemb}.
\textbf{Skip-thought}~\cite{Kiros:2015} is a sequence-to-sequence model that encodes sentences using a Recurrent Neural Network
(RNN) with Gated Recurrent Units (GRUs)~\cite{Cho:SSST:2014}. 
\textbf{FastSent}~\cite{Hill:NAACL:2016} is similar to \textbf{Skip-thought} in that both models predict 
the words in the next and previous sentences given the current sentence.
However, unlike \textbf{Skip-though} which considers the word-order in a sentence,
\textbf{FastSent} models a sentence as a bag-of-words. 
\textbf{Paragraph2Vec}~\cite{Le:ICML:2014} learns a vector for every short-text (eg. a sentence) in a corpus
jointly with word embeddings for every word in that corpus such that the word embeddings are shared across
all short-texts in the corpus.
Sequential Denoising Autoencoder (\textbf{SDAE})~\cite{Hill:NAACL:2016}
 is an encoder-decoder model with a Long Short-Term Memory (LSTM)~\cite{Hochreiter:1997} unit.
 We use the \textbf{SDAE} version that uses pre-trained CBOW embeddings to initialise the word embeddings 
 because of its superior performance over the \textbf{SDAE} version that uses randomly initialised word embeddings.
 
 We use Convolutional Neural Networks (\textbf{CNN}) for creating sentence-level embeddings as a baseline.
 For this purpose, we follow the model architecture proposed by \citet{kim:2014:EMNLP2014}.
 Specifically, each word $v_{i}$ in a sentence is represented by a $d$-dimensional word embedding $\vec{v}_{i} \in \R^{d}$, and the word embeddings are concatenated to create a fixed-length sentence embedding. The maximum length $n$ of a sentence is used to determine the length of this initial sentence-level embedding, where sentences with words less than this maximum length are padded using null vectors.
 Next, a convolution operator defined by a filter $\vec{w} \in \R^{hd}$ is applied on windows of consecutive $h$ tokens in sentences to produce new feature vectors for the sentences. 
 We use several convolutional filters by varying the window size.
 Next, max-over-time pooling~\cite{Collobert:2011} is applied on this feature map to select the maximum value corresponding to a particular feature. This operation produces a sentence-level embedding that is independent of the length of the sentence.
 Finally, a fully connected layer with dropout~\cite{Srivastava:2014} and a softmax output unit is applied on top of this sentence representation that can predict the class label of a sentence. Pre-trained CBOW embeddings are used in the CNN-based sentence encoder as well.

From Table~\ref{tbl:sentemb} we see that the proposed \textbf{Global Feature Expansion} method obtains 
best classification accuracies on \textbf{MR} and \textbf{CR} datasets with statistically significant improvements
over the corresponding second-best methods, whereas \textbf{Skip-thought} reports the best results on the \textbf{SUBJ} and \textbf{TREC} datasets.
However, unlike \textbf{Skip-thought} that is trained for two weeks on a GPU cluster, ClassiNets can be trained
in less than 6 hours end-to-end on a single core CPU.
The computational efficiency of ClassiNets is particularly attractive when continuously classifying large amounts of short-texts such as, for example,
sentiment classification of tweets coming in as a continuous data stream.

\begin{table}[t]
\caption{Comparison against sentence-level embedding methods.}
\begin{center}
\begin{tabular}{l c c c c} \toprule
Method					& \textbf{MR} 			& \textbf{CR}				&	\textbf{SUBJ} 		& \textbf{TREC}\\ \midrule
Skip-thought				& $76.5$				& $80.1$				&	$\mathbf{93.6}^{*}$ 	& $92.2$ \\	
Paragraph2Vec			& $74.8$				& $78.1$				&	$90.5$ 				& $59.4$ \\
FastSent					& $70.8$				& $78.4$				&	$88.7$  				& $76.8$ \\
SDAE					& $74.6$				& $78.0$				&	$90.8$ 				& $77.6$ \\
CNN						& $76.1$				& $79.8$				&	$89.6$ 				& $83.4$\\
Global Feature Expansion	& $\mathbf{81.2}^{*}$		& $\mathbf{83.89}^{*}$	&	$89.7$ 				& $88.3$ \\ \bottomrule
\end{tabular}
\label{tbl:sentemb}
\end{center}
\end{table}

\subsection{Qualitative evaluation}
\label{sec:quality}

\begin{table*}[t]
\caption{Example short-reviews and the features predicted by ClassiNet. The correct label (+/-) is shown
within brackets. All these instances were misclassified when classified using the original features.
However, when we use the features predicted by the ClassiNet all those instances are correctly classified.}
\begin{center}
\begin{tabular}{|p{7cm}|p{7cm}|} \hline
Review	& Predicted features \\ \hline \hline
On its own cinematic terms, it successfully showcases the passions of both the director and novelist Byatt. (+) & 
\emph{writer, played, excellent, thriller, story, writing, subject, script, animation, films, role, storyline, experience, episode, cinematography.} \\  \hline
What Jackson has accomplished here is amazing on a technical level. (+) & 
\emph{beautiful, perfect, fantastic, good, brilliant, great, wonderful, excellent, fine, strong.} \\ \hline
This is art playing homage to art. (+) &
\emph{cinema, modern, theme, theater, reality, style, experience, British, drama, documentary, history, period, acting, cinematography.} \\
\hline
About as satisfying and predictable as the fare at your local drive through. (-) &
\emph{terrible, ridiculous, annoying, least, horrible, poor, slow, awful, dull, scary, boring, stupid, bad, silly.} \\ \hline
\end{tabular}
\end{center}
\label{tbl:example}
\end{table*}%

In Table~\ref{tbl:example}, we show the expansion candidates predicted by the proposed \textbf{Global Feature Expansion} method for
some randomly selected short-reviews. The gold standard sentiment labels associated with each short review in the test dataset are shown
within brackets. All the reviews shown in Table~\ref{tbl:example} are misclassified if we had used only the
features in the original review. However, by appending the expansion features found from the ClassiNet,
we can correctly predict the sentiment for those short reviews. From Table~\ref{tbl:example}, we see that 
many semantically related features are found by the proposed method.

\begin{figure}[t]
\centering
\includegraphics[height=6cm]{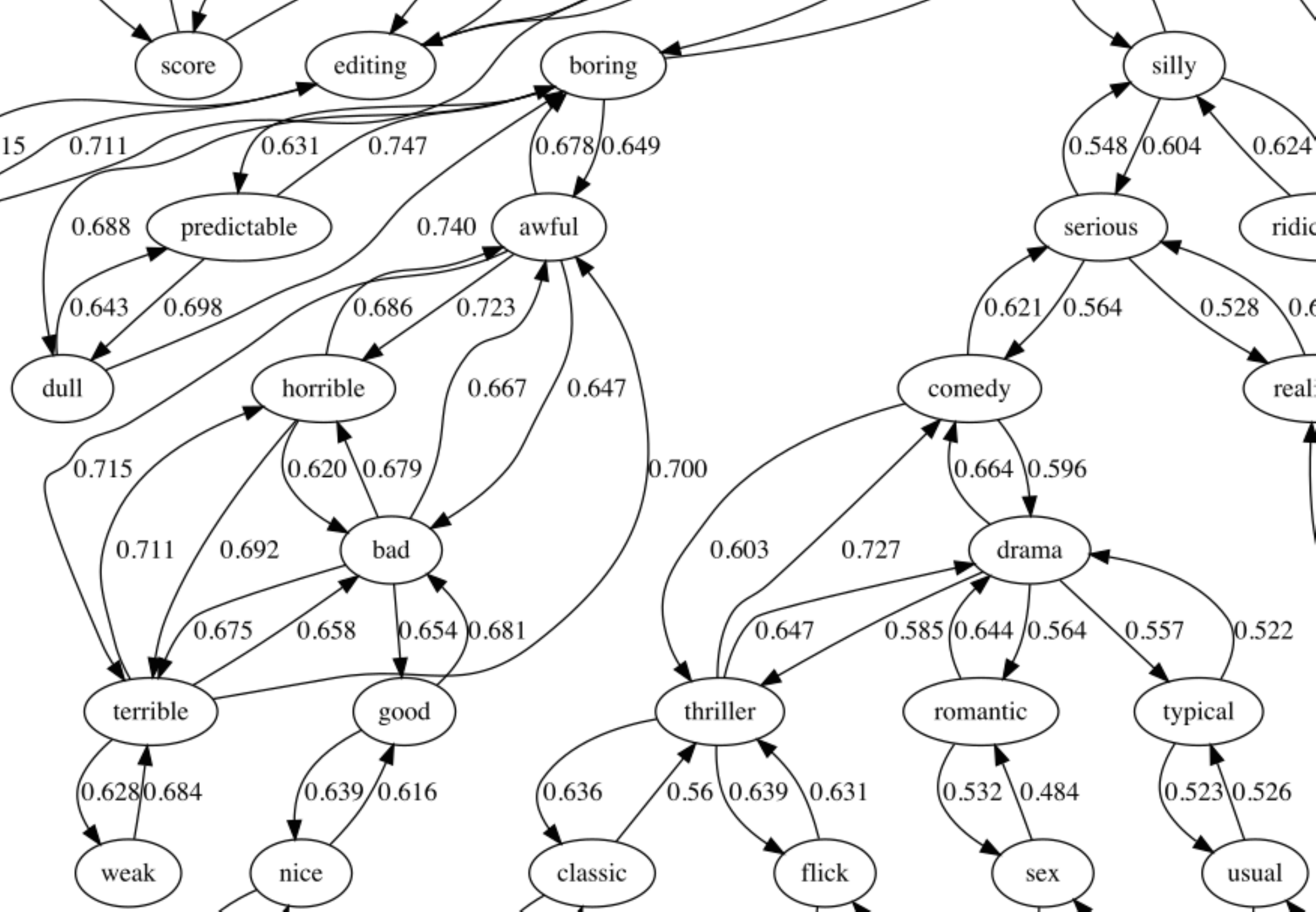}
\caption{Portion of the created ClassiNet from movie reviews. Vertices denote features and the edge-weights are shown on arrows.}
\label{fig:classinet}
\end{figure}

Figure~\ref{fig:classinet} shows an extract from the ClassiNet we create from the Large Movie Review dataset.
To avoid cluttering of edges, we show only the edges 
for a sparse $k=4$ mutual neighbour graph created from the original densely connected ClassiNet.
First, for each vertex $v_i$ in the ClassiNet we compute its top $k$ similar vertices according to the edge weights. 
Next, we connect a vertex $v_i$ to a vertex $v_j$ in the $k$-mutual neighbour graph if $v_j$ is among the top $k$ similar vertices of $v_i$,
and $v_i$ is among the top $k$ similar vertices of $v_j$. 
We see that synonyms, such as \emph{awful}, and \emph{horrible} are connected by high weighted edges in Figure~\ref{fig:classinet}.
It is interesting to see that antonyms, such as \emph{good}, and \emph{bad} are also among the mutual nearest neighbours because
those terms frequently occur in similar contexts (e.g., \emph{good movie} vs. \emph{bad movie}).
Moreover, Figure~\ref{fig:classinet} shows the importance of propagating over the ClassiNet, instead of simply considering
the directly connected vertices as the expansion candidates.
For example, although being highly related features, there is no direct connection from \emph{horrible} to \emph{boring}
in the ClassiNet. However, if we consider two-hop connections then we can find a path through \emph{awful}.

\subsection{Effect of the Damping Factor}
\label{sec:damp}
\begin{figure}[t]
\begin{center}
\includegraphics[height=6cm]{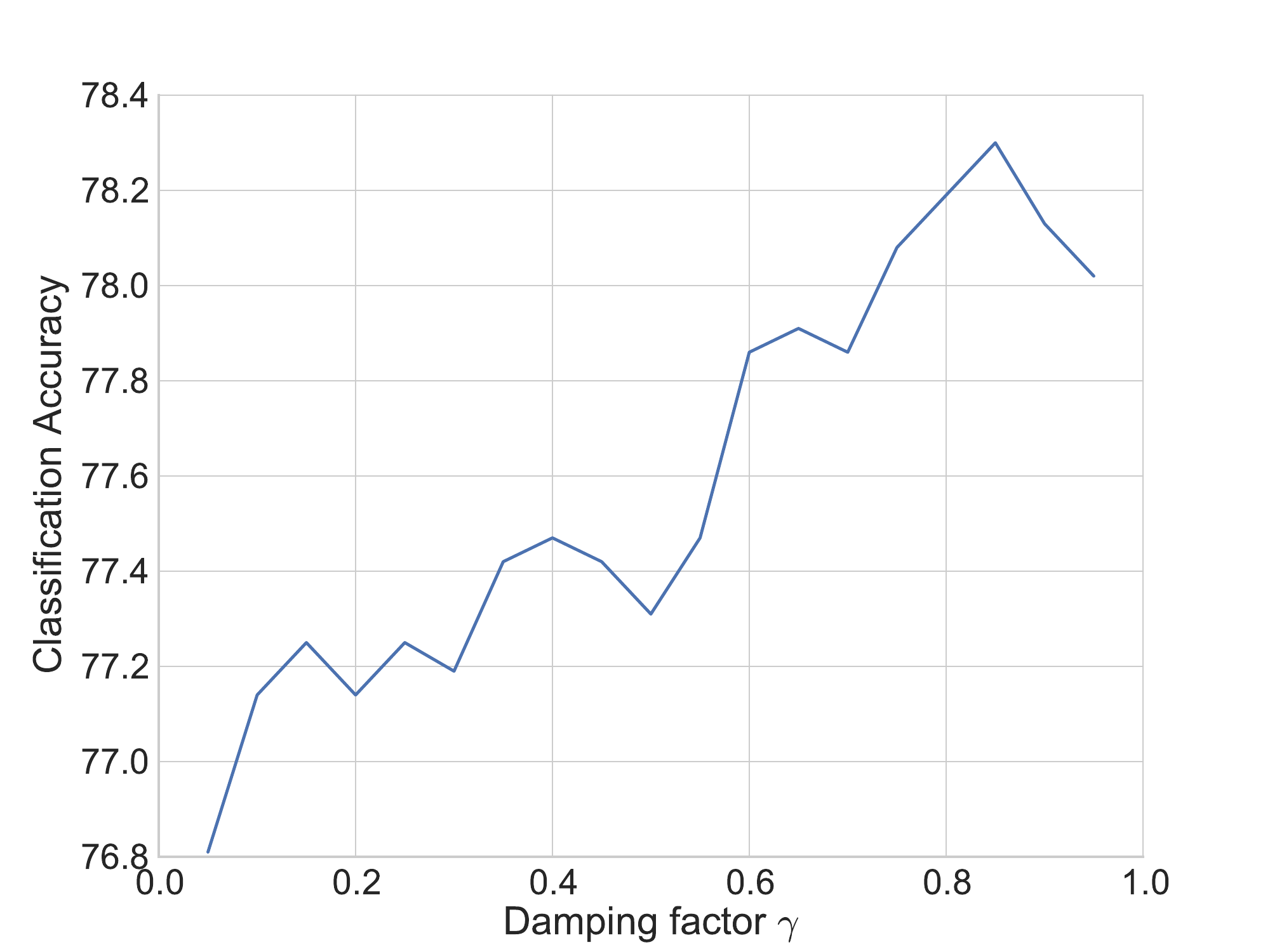}
\caption{The effect of the damping factor on the classification accuracy out.}
\label{fig:damp}
\end{center}
\end{figure}

To empirically study the effect of the damping factor on the classification accuracy of short-texts under the \textbf{Global Feature
Expansion} method, we randomly select $1000$ positive and $1000$ negative sentiment labeled
sentences from the Large Movie Review dataset as validation data, 
and evaluate the sentiment classification accuracy 
of the \textbf{Global Feature Expansion} method with different $\gamma$ values. The result is shown in Figure~\ref{fig:damp}.
Note that smaller $\gamma$ values will reduce the propagation than larger $\gamma$ values, restricting the expansion candidates
to a smaller local neighbourhood surrounding the original features.
From Figure~\ref{fig:damp} we see that initially when increasing $\gamma$ the classification accuracy increases and reaches a peak at
$\gamma = 0.85$. This shows that it is indeed important to find expansion neighbours by propagating over the ClassiNet as done by the
global feature expansion method. However, setting $\gamma > 0.85$ results in a drop of classification accuracy, which is due to
distant and potentially irrelevant expansion candidates.
Interestingly, $\gamma = 0.85$ has been found to be the optimal value for different graph-based propagation tasks such as 
the PageRank~\cite{PageRank}.

% discussion points:
% Short text classification is only one of the many applications of ClassiNets. We can use it for other feature suggesion
% tasks such as query suggestion, keyword suggestion, tag recommendation etc.
% This is a strength of explicit feature augmentation approaches such as ClassiNets that is not available with
% latent projection approaches such as skip-thought vectors, FastSent, paragraph2vec etc.
% If text classification was the sole purpose then we could use implicit or explicit feature augmentation approaches.
% This could be even mentioned when we introduce word2vec (skip-gram) baseline in the Experiments section.

\subsection{Number of Expansion Features}
\label{sec:featcount}

In this Section we analyse the number of feature appended to train/test instances by the different feature expansion methods using a fixed ClassiNet.
Recall that none of the feature expansion methods we proposed has any predefined number of expansion features.
In contrast, the number of expansion features depends on several factors: (a) the number of features in the original (prior to expansion) feature vector, (b) the size and the connectivity of the ClassiNet and (c) the feature expansion method. For example, if a particular feature vector has $n$ features, which are all present in the ClassiNet, then on average under the All Neighbour Expansion method, we will append $dn$ number of features to this instance where $d$ is the out degree of the ClassiNet. More precisely, the actual number of expansion features will be different from $dn$ due to several reasons. First, some vertices in ClassiNet might have different numbers of neighbours, not necessarily equal to the out degree. Second, the out degree considers the weight of the edges and not simply the different number of vertices connected via outbound edges. Third, some of the expansion features might already be in the original feature vector, thereby not increasing the number of features. Finally, the same expansion feature might be suggested by different vertices, therefore doubly counting the number of expansion features. 

To empirically analyse the number of expansion features, we build a ClassiNet containing 700 vertices and count the number of features expanded on the \textbf{SUBJ} train dataset. The out degree $d$ is given by \eqref{eq:out-degree}.
\begin{equation}
 \label{eq:out-degree}
 d = \frac{1}{N} \sum_{i} \sum_{j \in \cN(v_{i})} w_{ij}
\end{equation}
Here, $N$ is the total number of vertices in the ClassiNet, $\cN(v_{i})$ is the set of neighbours connected to $v_{i}$ by an out bound link, and $w_{ij}$ is the weight of the edge connecting vertex $v_{i}$ to $v_{j}$.

\begin{figure}[t]
\begin{center}
\includegraphics[height=6cm]{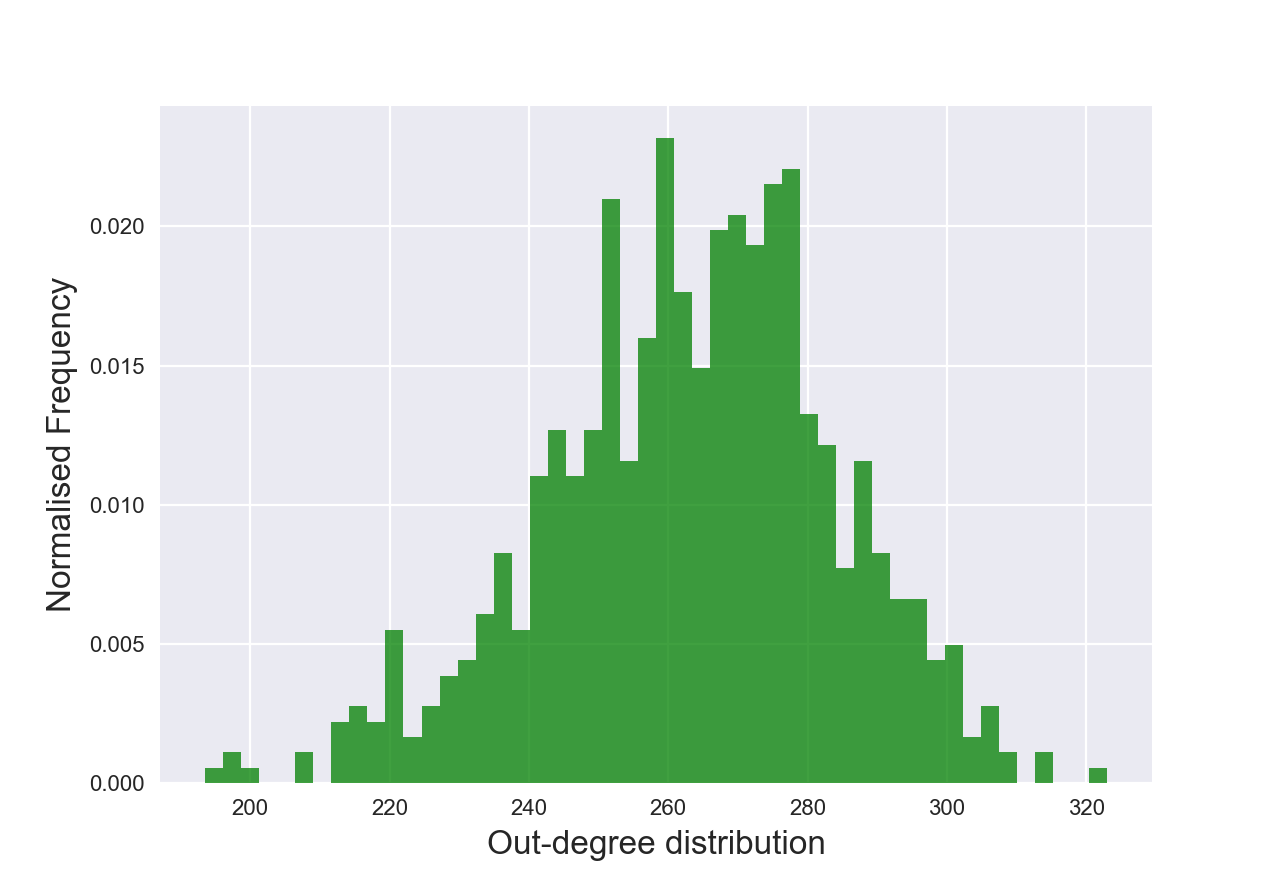}
\caption{Out degree distribution of the ClassiNet.}
\label{fig:degree}
\end{center}
\end{figure}

\begin{figure}[t]
\begin{center}
\includegraphics[height=6cm]{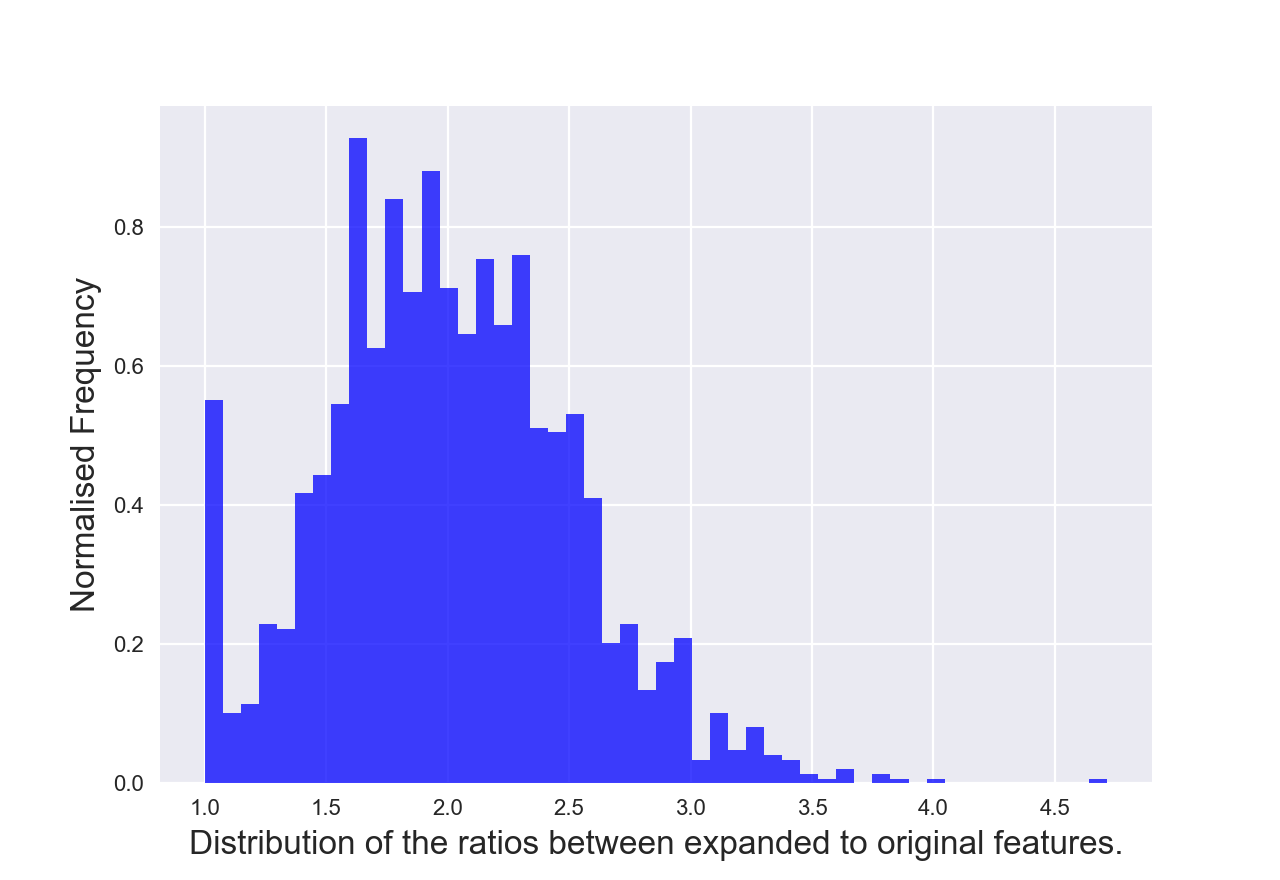}
\caption{All neighbour Expansion.}
\label{fig:all-neighb}
\end{center}
\end{figure}

Figure~\ref{fig:degree} shows the degree distribution for the ClassiNet with degree $d = 263.35$.
We see that most vertices are connected to $240-300$ other vertices in the ClassiNet. Given that this ClassiNet contains 700 vertices, this is a tightly connected, dense graph. For each train instance in the \textbf{SUBJ} dataset, we compute the expansion ration, ratio between the number of features after and before feature expansion, for the All Neighbour Expansion (Figure~\ref{fig:all-neighb}) and Global Feature Expansion (Figure~\ref{fig:global}).
We see that the expansion ratio is higher for the global feature expansion (ca. 25-30) compared to that for all neighbour expansion (ca. 1.5-2.5).
Given that the global feature expansion considers a broader neighbourhood surrounding the initial features in an instance this is not surprising. Moreover, it provides an explanation for the superior performance of the global feature expansion. 
Although expanding too much using not only relevant nearby features but also potentially irrelevant broader neighbourhoods is likely to degrade performance, we see that at the level of expansions done by the global feature expansion this is not an issue.
Therefore, we conclude that under the global feature expansion method, we do not need to impose any predefined limitations to the number of expansion features. 

\begin{figure}[t]
\begin{center}
\includegraphics[height=6cm]{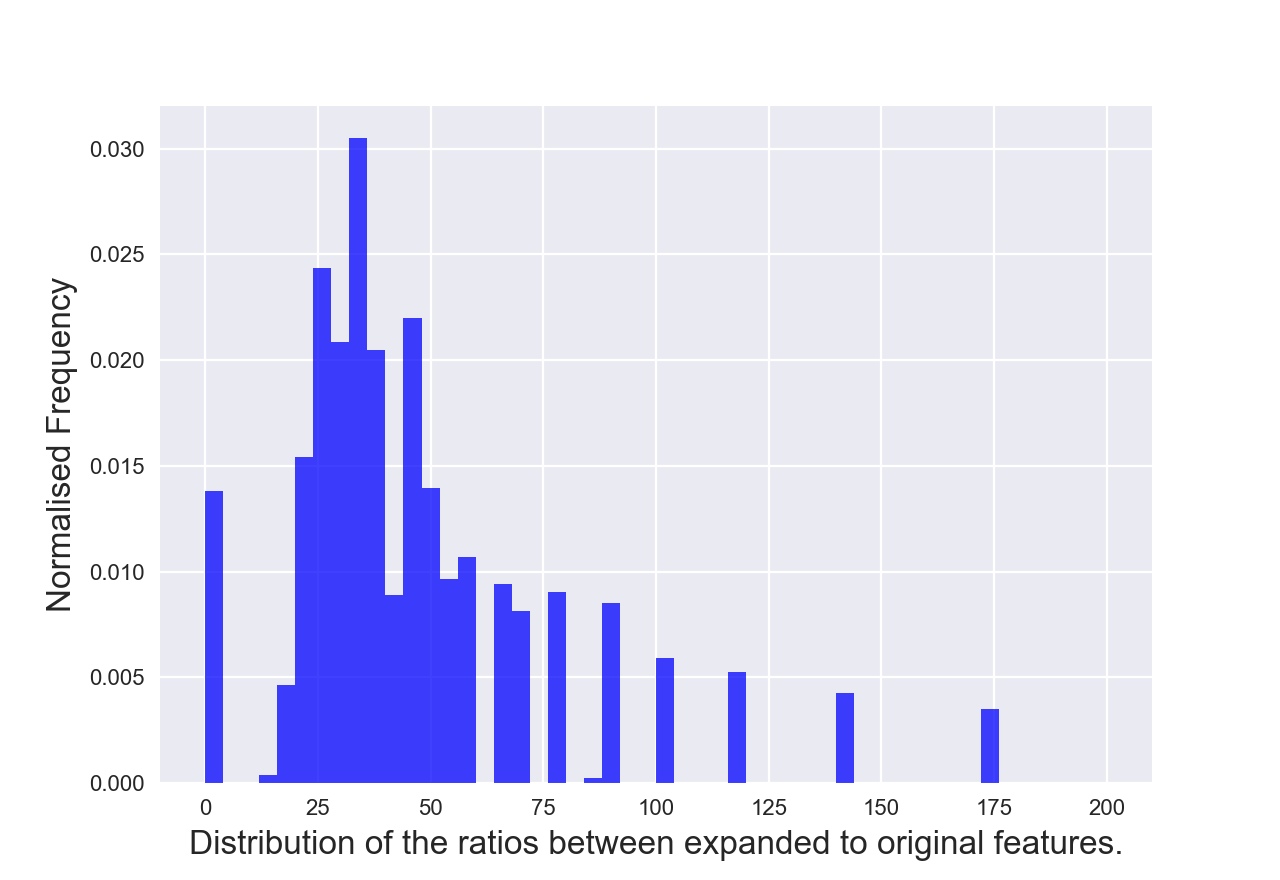}
\caption{Global Feature Expansion.}
\label{fig:global}
\end{center}
\end{figure}

\section{Conclusion}
\label{sec:conclusion}

We proposed ClassiNet, a network of binary classifiers for predicting missing features to overcome the feature sparseness
problem observed in short-text classification. We select positive and negative training instances for learning the
feature predictors using unlabeled data. 
In ClassiNets, the weight of the edge connecting the vertex $v_i$ to $v_j$ represents the probability that
given $v_i$ is predicted to occur in an instance, $v_j$ is also predicted to occur in the same instance.
We proposed an efficient method using locality sensitive hashing to approximately compute the neighbourhood of a vertex,
thereby avoiding all-pair computation of confusion matrices. We propose local and global methods for feature expansion
using ClassiNets. Our experimental results show that the global
feature expansion method significantly improves the classification accuracy of a sentence-level sentiment
classification tasks outperforming previously proposed methods such as structural correspondence learning (SCL),
and frequent term sets (FTS), Skip-thought vectors, FastSent, and Paragraph2Vec on multiple datasets. 
Moreover, close inspection of the expanded feature vectors show that features that are related to
an instance are found as expansion candidates for that instance. In the future, we plan to apply ClassiNets to other tasks that
require missing feature prediction such as recommendation systems.

\bibliographystyle{ACM-Reference-Format}
\bibliography{ClassiNet}

\end{document}